\documentclass{article} 
\usepackage{iclr2027_conference,times}

\usepackage{amsmath,amsfonts,bm}

\def\eqref#1{equation~\ref{#1}}

\def\1{\bm{1}}

\DeclareMathAlphabet{\mathsfit}{\encodingdefault}{\sfdefault}{m}{sl}
\SetMathAlphabet{\mathsfit}{bold}{\encodingdefault}{\sfdefault}{bx}{n}

\usepackage{hyperref}
\usepackage{url}
\usepackage{hyperref}
\usepackage{booktabs}
\usepackage{multirow}
\usepackage{amsmath}
\usepackage{amssymb}
\usepackage{bbding}
\usepackage{arydshln}
\usepackage{xcolor}
\usepackage{graphicx}
\usepackage{wrapfig}
\usepackage{enumitem}
\usepackage{tabularx}
\usepackage{makecell}
\usepackage{array}
\usepackage{subcaption}
\usepackage{wrapfig}
\usepackage{tabularx}
\usepackage{inconsolata} 
\usepackage{enumitem}
\usepackage{ragged2e}
\usepackage{svg}
\usepackage[most]{tcolorbox}
\newtcolorbox{promptbox}[1]{
  enhanced,
  breakable,
  colback=white,
  colframe=black,
  boxrule=1pt,
  arc=3pt,
  left=10pt,
  right=10pt,
  top=8pt,
  bottom=8pt,
  title={#1},
  coltitle=white,
  colbacktitle=black,
  fonttitle=\bfseries\rmfamily,
  boxed title style={
    colback=black,
    colframe=black,
    sharp corners
  }
}

\usepackage{xcolor}
\definecolor{ao}{rgb}{0.0, 0.5, 0.0}
\definecolor{forestgreen}{RGB}{190, 232, 194} 
\definecolor{brightyellow}{RGB}{254, 241, 138} 
\usepackage{dashrule}

\title{The Anatomy of Socio-Demographics in Pluralistic Offensiveness Modeling: When and Where It Helps, and the Role of Reasoning}

\author{\textbf{Yifan Wang\textsuperscript{1},}
        \textbf{Leixin Zhang\textsuperscript{2},}
        \textbf{Jinyi Mu\textsuperscript{3},}
        \textbf{Yu Wang\textsuperscript{4},}
        \textbf{Mayank Jobanputra\textsuperscript{1},} 
        \\
        \textbf{Isabel Valera\textsuperscript{1,5},}
        \textbf{Vera Demberg\textsuperscript{1,6}} 
        \vspace{0.6em}
        \\
        \textsuperscript{1} Saarland University  \space \textsuperscript{2} University of Tübingen
        \textsuperscript{3} University of California San Diego 
        \vspace{0.6em}
        \\
        \textsuperscript{4} Utrecht University
        \textsuperscript{5} Max Planck Institute for Software Systems 
        \vspace{0.6em}
        \\
        \textsuperscript{6} Max Planck Institute for Informatics \vspace{0.4em} \\
        \texttt{yifwang@lst.uni-saarland.de}
}

\iclrfinalcopy 
\begin{document}

\maketitle
\fancyhead{}

\begin{abstract}
Subjective judgments such as offensiveness often vary substantially across individuals, motivating pluralistic modeling approaches that incorporate annotators' socio-demographic information to capture this variation.
However, prior findings on the benefits of socio-demographic information for pluralistic modeling are mixed, calling for a systematic analysis of both its effectiveness and the mechanisms through which it improves prediction.
In this work, we study \textbf{when} and \textbf{where} socio-demographics benefit pluralistic offensiveness modeling, as well as \textbf{the role of explicit reasoning} in leveraging this information.
Across three multi-annotator datasets with prompted and trained LLMs, we first show that demographic gains are small and inconsistent for individual predictions but become clearer at the group and population levels, because aggregation attenuates the idiosyncratic variation that masks small but systematic demographic signals.
Second, using a reasoning and prediction format that explicitly decomposes individual judgment into an overall baseline and a demographic-specific deviation, we show that the gains arise from both components: demographic information improves baseline estimation and enables meaningful demographic-specific shifts, and interventions show that better component estimates improve final predictions.
Finally, comparing direct demographic conditioning with explicit reasoning about how the text interacts with annotator demographics, we find that reasoning yields more accurate baseline and shift estimates as well as final predictions, whereas conditioning alone rarely produces informative shifts.
These results suggest that the mixed evidence in prior work may partly stem from individual-level evaluation, and that the value of socio-demographic information depends on both how it is evaluated and how models are guided to use it.

\end{abstract}

\section{Introduction}

Subjective judgments such as sentiment, bias, and offensiveness often vary substantially across annotators.
Rather than treating such disagreement as noise, pluralistic modeling aims to preserve and model the diversity of human judgments~\citep{pmlr-v235-sorensen24a,davani-etal-2022-dealing, fleisig-etal-2023-majority}.
A common approach is to incorporate annotators' socio-demographic information, motivated by the observation that social and cultural backgrounds can systematically shape how the same text is perceived.
However, prior studies provide mixed evidence on its benefits for pluralistic modeling: demographic information improves prediction in some settings or along particular demographic axes, while providing limited benefits or even degrading performance in others~\citep{mokhberian-etal-2024-capturing, zhang-coltekin-2026-modeling, orlikowski-etal-2023-ecological, orlikowski-etal-2025-beyond}.
These conflicting findings call for a systematic analysis of when socio-demographic information is beneficial, where its benefits arise, and how modeling choices affect its use.\footnote{A more detailed discussion of related work is provided in Appendix~\ref{app:related work}.}

We first ask \textbf{when} socio-demographic information is beneficial across three levels of evaluation: individual annotators, socio-demographic groups, and the overall population.
Existing work has evaluated demographic-aware modeling at these levels, but largely treats them as separate targets rather than examining how the benefits change across them~\citep{fleisig-etal-2023-majority, parappan-henao-2025-learning, xu-etal-2026-modeling}.
This gap can obscure small but systematic demographic signals that are difficult to detect at the individual level yet become apparent after group or population aggregation.
We find that demographic gains are indeed limited at the individual level and become more pronounced at the group and population levels.
A simple statistical analysis explains why: aggregation attenuates residual individual variation, which demographics cannot explain, while preserving systematic demographic effects.

\begin{figure}[t]
    \centering
    \includegraphics[width=\textwidth]{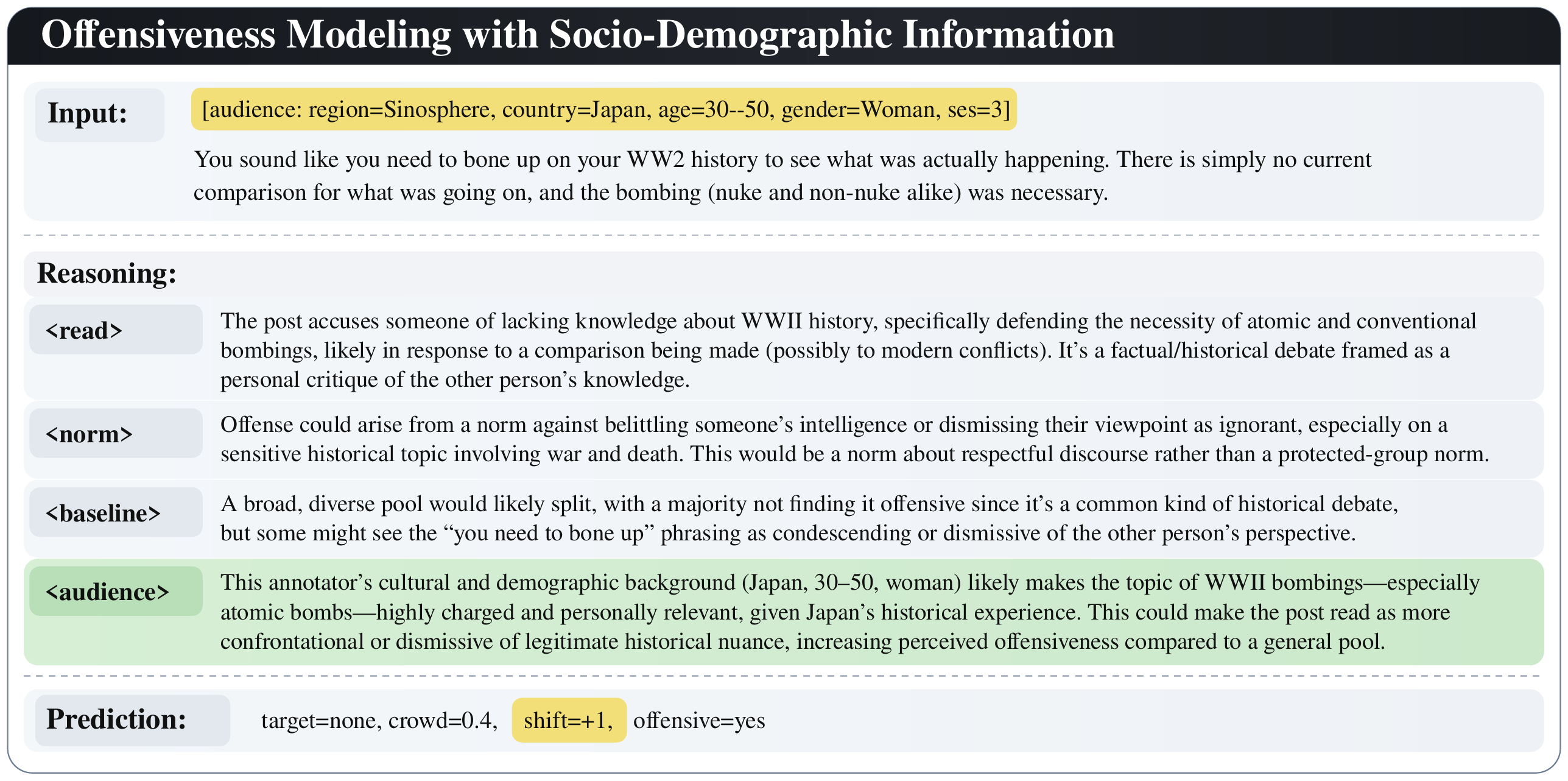}
    \caption{Example from D3CODE using the trained Qwen-4B model.
Variants differ in whether they incorporate the annotator's {\setlength{\fboxsep}{0.5pt}\colorbox{brightyellow}{socio-demographic profile}} and the {\setlength{\fboxsep}{0.5pt}\colorbox{brightyellow}{demographic-specific deviation}}, and {\setlength{\fboxsep}{0.5pt}\colorbox{forestgreen}{socio-demographics-aware reasoning}}.
Here, the model correctly estimates a non-offensive baseline and, given the historical sensitivity of the topic, a positive shift for the Japanese annotator, leading to the correct offensive individual prediction despite the non-offensive majority judgment.}
    \label{fig:reasoning_example}
\vspace{-5mm}
\end{figure}

We next ask \textbf{where} in the prediction process these benefits arise.
An individual judgment can be viewed as an overall baseline plus a demographic-specific deviation, but existing approaches leave this decomposition implicit and expose only the final prediction~\citep{xu-etal-2026-modeling, zhang-coltekin-2026-modeling}.
It is therefore unclear whether socio-demographics help by improving the baseline, the demographic-specific deviation, or both.
We instead use a reasoning and prediction format that exposes both estimates: as illustrated in Figure~\ref{fig:reasoning_example}, the model first estimates how a broad population would perceive the text (\texttt{crowd}) and then whether the target annotator would deviate from this baseline (\texttt{shift}) before making its final prediction.
This decomposition makes each component inspectable: in the example, we can verify that the model correctly estimates a non-offensive \texttt{crowd} baseline and a positive \texttt{shift} for the Japanese annotator, reflecting the historical sensitivity of the topic, and thereby see why it makes the correct final offensive prediction despite the non-offensive majority.
We find that socio-demographic information improves both components, and that interventions on them change final predictions in the expected direction.

Finally, we study \textbf{the role of reasoning} in leveraging demographic information.
Throughout our analyses, we compare direct demographic conditioning with a variant that explicitly reasons about how the text interacts with the annotator's demographics.
We find that explicit reasoning yields more accurate baseline and shift estimates than conditioning alone, which in turn improve aggregated predictions.
Reasoning thus does not merely improve the final prediction; it strengthens how demographic information is used within each component.

In summary, our contributions are:
(1) a multi-level evaluation showing that the value of socio-demographic information is underestimated by individual-level metrics, together with a statistical account of why;
(2) a component-level analysis, enabled by explicitly decomposing predictions into a baseline and a demographic-specific shift, showing through evaluation and interventions that demographic gains arise from both improved baseline estimation and meaningful deviation estimation; and
(3) a controlled comparison of demographic conditioning and demographic reasoning across three datasets and both prompted and trained LLMs, showing that explicit reasoning is key to effectively using socio-demographic information, yielding more accurate component estimates and improved predictions.

\section{Incorporating Socio-Demographics in Offensiveness Modeling}
\label{sec:model}

We describe the task, the decomposition underlying our analyses, the model variants, and the experimental setup shared by Sections~\ref{sec:when} and~\ref{sec:where}.

\paragraph{Problem Formulation}
Consider a dataset $\mathcal D=\{(x_i,a_i^j,y_i^j)\}$, where $x_i\in\mathcal X$ is the text of item $i$, $a_i^j\in\mathcal A_i\subseteq\mathcal A$ is the $j$-th annotator assigned to item $i$, and $y_i^j\in\{0,1\}$ indicates whether annotator $j$ labels $x_i$ as offensive. Here, $\mathcal A$ denotes the annotator space of the dataset, and $\mathcal A_i$ the set of annotators of item $i$.
Each annotator $a_i^j$ is associated with a multi-attribute socio-demographic profile $\mathbf z_i^j$.
We use $g\in\mathcal G$ to denote a demographic group, i.e., a particular value along a demographic axis.

An individual judgment can be decomposed into three components: an item-level baseline capturing how a broad pool of annotators would perceive the text, a demographic-specific deviation capturing how the annotator's demographic profile systematically shifts this perception for the item, and a zero-mean residual capturing idiosyncratic differences not explained by the profile:
\begin{equation}
\underbrace{y_i^j}_{\text{individual}} \;=\; \underbrace{c^\star(x_i)}_{\text{baseline}} \;+\; \underbrace{\delta^\star(x_i, \mathbf z_i^j)}_{\text{demographic-specific deviation}} \;+\; \underbrace{\varepsilon_i^j}_{\text{zero-mean residual}},
\label{eq:decomp}
\end{equation}
Existing approaches model $y_i^j$ directly, leaving the decomposition implicit. If a model instead outputs estimates of the baseline and the deviation, we can evaluate not only its final predictions $\hat{y}_i^j$ against $y_i^j$, but also the effect of demographic information on each component separately.


\paragraph{Variants of Reasoning and Prediction Format}

We therefore prompt or train LLMs to follow the reasoning and prediction format in Figure~\ref{fig:reasoning_example}, which emits the baseline, the demographic-specific shift, and the final individual prediction as separate outputs.
We consider three variants that differ in whether and how they use annotators' socio-demographic information.
The basic variant, \textbf{NoDemo}, is annotator-blind and receives only the post.
It reasons about the content through \texttt{read}, which summarizes the post, \texttt{norm}, which identifies potentially violated social norms, and \texttt{baseline}, which estimates how a broad population would perceive it.
It then predicts the relevant \texttt{target}, the baseline (\texttt{crowd}$\in[0,1]$), and the binary individual judgment (\texttt{offensive}$\in\{\texttt{yes},\texttt{no}\}$).
\textbf{DemoCond} (demographic conditioning) additionally receives the annotator's socio-demographic profile and predicts a demographic-specific \texttt{shift}$\in\{-1,0,+1\}$ relative to the baseline, but does not explicitly reason about the effect of demographics.
\textbf{DemoReason} (demographic reasoning) further adds an \texttt{audience} reasoning block that explicitly reasons about how the target demographic features may shift the baseline before producing the \texttt{shift} and \texttt{offensive} prediction.\footnote{We encode only the direction of the demographic effect in \texttt{shift}, as magnitude estimation is unstable and prone to collapse. Zero-shot and free-form CoT baselines are reported in Appendix~\ref{app:full_performance}.}

\paragraph{Prompting and Training}

We evaluate both prompted and trained LLMs. 
In the prompting setting, LLMs receive task instructions together with the reasoning and prediction format of each variant.
For training, we adopt a standard two-stage post-training pipeline~\citep{fan-etal-2025-biasguard, ICLR2026_8e3b8de2}. 
Claude Sonnet 5 first generates cold-start reasoning trajectories for each variant using its corresponding inputs, the gold individual \texttt{offensive} label, and the empirical item-level mean of individual annotations as the \texttt{crowd} target.
To avoid leaking demographic information into the reasoning, DemoCond reuses the reasoning blocks generated for NoDemo (without access to demographics), paired with the prediction line of the corresponding DemoReason trajectory.
We then perform SFT on the resulting trajectories, followed by GRPO with rewards based on the correctness of the final individual predictions.

\paragraph{Datasets and Models}
We conduct experiments on three multi-annotator corpora: DICES~\citep{dices-2023}, D3CODE~\citep{mostafazadeh-davani-etal-2024-d3code}, and POPQUORN~\citep{pei-jurgens-2023-popquorn}.
Each dataset provides multiple annotations per item (81.0/32.6/8.7 on average), enabling empirical estimates of overall baselines and proxy targets for demographic-specific deviations.
Table~\ref{tab:datasets} in Appendix~\ref{app:details} summarizes the dataset statistics and socio-demographic coverage.
We train four open-source Llama-3.x and Qwen-3.5 models ranging from 3B to 9B parameters, and prompt nine models: open-source Llama and Qwen models ranging from 3B to 122B, and two commercial models, GPT-4o-mini and Haiku-4.5.
Model configurations, training and prompting details, and additional baselines are provided in Appendix~\ref{app:details}.

\section{When Do Socio-Demographics Help? The Role of Aggregation}
\label{sec:when}

We first ask whether the benefits of incorporating annotator socio-demographics differ across levels of aggregation: individual, group, and population.

\paragraph{Evaluation Setup}

We evaluate model predictions at the individual, group, and population levels.
At the individual level, each prediction is compared directly with its corresponding annotation.
At the group level, model predictions and annotations are aggregated within each item--group pair to obtain model-estimated and empirical judgments for a specific socio-demographic group.
At the population level, predictions are aggregated across all annotators for each item to obtain model-estimated and empirical population judgments.\footnote{Note that the model-emitted \texttt{crowd} baseline and aggregated individual predictions both estimate broad population perception but are computed differently.
We refer to the former as baseline or \texttt{crowd} baseline, and to the latter as population or population-level judgments.
Their modeling performance is compared in Table~\ref{tab:crowd}.}

We then evaluate these predictions using both discrete and continuous metrics.
For discrete evaluation, the individual predictions are evaluated directly, while group- and population-level estimates are binarized.
We report F1 score (\textbf{F1}) and balanced accuracy (\textbf{Acc}).
For continuous evaluation at the group and population levels, we directly compare the aggregated model estimates with their empirical targets using mean squared error (\textbf{MSE}) and Spearman correlation (\textbf{$\rho$}).

\begin{table}[h]
\centering
\small
\caption{
        Individual-, group- and population-level results for prompted and trained models. 
        Scores are averaged across all dataset--model pairs. 
        * / ** indicate significant improvement over NoDemo / over both NoDemo and the other demographic-aware variant (paired $t$-test, $p<0.05$). 
        \textbf{The benefits of incorporating socio-demographics become more apparent at higher levels of aggregation.}
        }
\resizebox{\textwidth}{!}{%
\begin{tabular}{llcccccccccc}
\toprule
\multirow{2}{*}{\textbf{Setting}} & \multirow{2}{*}{\textbf{Variant}} & \multicolumn{2}{c}{\textbf{Individual}} & \multicolumn{4}{c}{\textbf{Group}} & \multicolumn{4}{c}{\textbf{Population}} \\
\cmidrule(lr){3-4}
\cmidrule(lr){5-8}
\cmidrule(lr){9-12}
 &  & F1 $\uparrow$ & Acc $\uparrow$ & F1$\uparrow$ & Acc$\uparrow$ & MSE$\downarrow$ & $\rho\uparrow$ & F1 $\uparrow$ & Acc $\uparrow$ & MSE $\downarrow$ & $\rho \uparrow$ \\
\midrule

  \multirow{3}{*}{Prompted} & NoDemo & 0.410 & 0.587 & 0.382 & 0.630 & 0.263 & 0.269 & 0.384 & 0.673 & 0.191 & 0.434 \\
 & DemoCond & 0.414 & \textbf{0.589} & 0.390 & 0.636 & \textbf{0.230*} & 0.295* & 0.389 & 0.680* & 0.132* & 0.526* \\
 & DemoReason & \textbf{0.426**} & 0.588 & \textbf{0.394*} & \textbf{0.639} & 0.235* & \textbf{0.301*} & \textbf{0.394*} & \textbf{0.690**} & \textbf{0.128*} & \textbf{0.543**} \\
 \midrule
\multirow{3}{*}{Trained} & NoDemo & \textbf{0.357} & \textbf{0.583} & 0.375 & \textbf{0.631} & 0.198 & 0.291 & 0.412 & 0.684 & 0.120 & 0.473 \\
 & DemoCond & 0.355 & 0.580 & \textbf{0.375} & 0.629 & \textbf{0.179*} & \textbf{0.303*} & 0.411 & 0.673 & 0.095* & 0.512* \\
 & DemoReason & 0.351 & 0.565 & 0.375 & 0.628 & 0.181* & 0.298* & \textbf{0.421*} & \textbf{0.686} & \textbf{0.075**} & \textbf{0.519*} \\

\bottomrule
\end{tabular}
}
\label{tab:performance}
\end{table}

\subsection{Socio-Demographic Gains Become More Visible with Aggregation}

Table~\ref{tab:performance} compares annotator-blind and demographic-aware variants and shows that the benefits of demographic modeling become increasingly visible at higher levels of aggregation.

At the individual level, demographic information provides little consistent advantage.
DemoCond and DemoReason achieve balanced accuracy comparable to NoDemo for prompted models, while both underperform NoDemo in F1 and Acc after training.
At the group level, the gains of demographic modeling become clearer.
On discrete metrics, demographic-aware variants begin to outperform NoDemo for prompted models and perform comparably to NoDemo after training, while significantly improving continuous metrics in both settings.

The advantage is strongest at the population level.
For prompted models, the F1 gains of DemoCond/DemoReason persist, while their Acc gains increase from 0.006/0.009 at the group level to 0.007/0.017, with DemoReason significantly outperforming NoDemo on both metrics.
For trained models, DemoReason also surpasses NoDemo in both F1 and Acc.
Continuous metrics show an even clearer aggregation effect. 
For prompted models, the MSE gains of DemoCond/DemoReason at the population level increase to approximately 1.8$\times$/2.3$\times$ their group-level gains, while the corresponding $\rho$ gains increase to 3.5$\times$/3.4$\times$.
Similarly, for trained models, the corresponding MSE gains at the population level increase by approximately 1.3$\times$/2.6$\times$, while the $\rho$ gains increase to 3.3$\times$/6.6$\times$ their group-level improvements.

Taken together, socio-demographic information captures useful systematic variation whose contribution is difficult to observe at the individual level but becomes increasingly visible as aggregation attenuates residual variation.
This pattern is particularly clear for the continuous metrics, which avoid information loss from binarization.
Similar patterns are observed across the other baselines and when transferring to unseen datasets.
Complete per-model results are reported in Appendix~\ref{app:full_performance}, with cross-dataset and held-out demographic-group generalization results in Appendix~\ref{app:generalization}.

Compared with simple demographic conditioning, explicit demographic reasoning generally yields larger gains, particularly at the group and population levels. 
Accordingly, DemoReason typically outperforms DemoCond at these levels, suggesting that explicit reasoning helps LLMs better exploit the systematic information conveyed by socio-demographic profiles.

\subsection{Aggregation Reveals Demographic Signal Obscured by Individual Noise}
\label{sec:residual dominance}

Why are demographic gains hard to observe individually but clear after aggregation?
Recall Eq.~\ref{eq:decomp}: $y_i^j = c^\star(x_i) + \delta^\star(x_i,\mathbf z_i^j) + \varepsilon_i^j$, with $\mathbb E[\varepsilon_i^j\mid x_i,\mathbf z_i^j]=0$. 
Let $p_i^j := c^\star(x_i)+\delta^\star(x_i,\mathbf z_i^j) = \mathbb E[y_i^j\mid x_i,\mathbf z_i^j]$ denote the systematic component, such that $y_i^j = p_i^j+\varepsilon_i^j$.
Socio-demographic information may improve estimation of $p_i^j$, but cannot explain $\varepsilon_i^j$.
Assume that the residuals are conditionally independent with a common variance $\sigma^2$ and independent of the model predictions.
When estimating the mean judgment $\bar y_i$ of $n$ annotators with the mean prediction $\bar{\hat y}_i$, the expected squared loss decomposes into a reducible model error and an irreducible residual variance term:
\begin{equation}
\mathbb E[(\bar{\hat{y}}_i-\bar y_i)^2]
=
\underbrace{\mathbb E[(\bar{\hat{y}}_i-\bar p_i)^2]}_{\text{reducible}}
+
\underbrace{\mathrm{Var}(\bar\varepsilon_i)}_{\text{irreducible}}
=
\underbrace{\mathbb E[(\bar{\hat{y}}_i-\bar p_i)^2]}_{\text{reducible}}
+
\underbrace{\sigma^2 / n}_{\text{irreducible}}.
\end{equation}
In subjective tasks, the residual captures idiosyncratic preferences and other unsystematic variation, which can constitute a substantial portion of prediction error.
Thus, at the individual level ($n=1$), residual variation can mask improvements in the systematic component.
From individual to group and population levels, the residual variance decreases at rate $1/n$, while systematic differences in expected judgments are preserved.
Aggregation therefore does not make socio-demographic information more informative; it makes its contribution easier to observe.

\paragraph{Empirical Evidence}
Two empirical observations support this explanation.
First, a bias-corrected decomposition of within-item label variation shows that no single demographic axis explains more than 6.2\% of the variation across datasets (D3CODE: $\leq6.2\%$; DICES: $\leq1.7\%$; POPQUORN: $\leq2.6\%$), indicating that demographics capture only a small systematic component of individual variation~\citep{hu-collier-2024-quantifying, sun-etal-2025-sociodemographic}.
Second, inter-annotator agreement differs little between annotator pairs from the same versus different demographic groups: the largest gap on the test split is only 3.1 percentage points, compared with overall agreement rates of 63--71\%.
Demographic effects are thus systematic but small relative to residual individual variation.

\vspace{-0.5em}
\begin{center}
\colorbox{gray!15}{%
  \parbox{0.98\linewidth}{%
    \textbf{Takeaway}: Socio-demographic information improves pluralistic offensiveness modeling, but its benefits are obscured by residual variation at the individual level and emerge clearly at the group and population levels. Explicit reasoning yields larger gains than simple conditioning.}}
\end{center}
\vspace{-0.5em}

\section{Where Do Socio-Demographics Help in the Prediction Process?}
\label{sec:where}

Leveraging the decomposition of individual predictions into a baseline and a demographic-specific shift, we next examine whether socio-demographics improve baseline estimation (Sections~\ref{sec:baseline_eval}--\ref{sec:baseline_mech}), demographic-specific shift estimation (Section~\ref{sec:shift_eval}), or both, and whether these improvements carry over to final predictions (Section~\ref{sec:intervention}).
We provide a qualitative comparison of the variants in terms of both prediction components and final predictions in Appendix~\ref{app:qualitative}. 

\subsection{Incorporating Socio-Demographics Improves Baseline Estimation}
\label{sec:baseline_eval}

Somewhat surprisingly, incorporating socio-demographics leads to more accurate estimates of the baseline, a quantity that should not depend on the target annotator.
We evaluate the model-emitted \texttt{crowd} against an empirical baseline $c(x_i)$, computed by averaging the individual annotations for item $i$.
Since these annotations come from demographically diverse annotator pools, The empirical baseline $c(x_i)$ should reliably approximate the underlying $c^\star(x_i)$.
Because demographic-aware variants may emit different \texttt{crowd} values for different annotators of the same item, we evaluate them at two levels: directly at the annotation level and, after averaging across annotators, at the item level.
We report mean squared error (\textbf{MSE}) and mean absolute error (\textbf{MAE}) for point-wise quality, together with Spearman correlation (\textbf{$\rho$}) and Wasserstein-1 distance (\textbf{W1}) for recovering the relative ordering and overall distribution of empirical baselines, respectively.

\begin{table}[h]
\centering
\small
\caption{
        Annotation- and item-level baseline estimation results for prompted and trained models.
        Scores are averaged across all dataset--model pairs.
        * / ** indicate significant improvement over NoDemo / over both NoDemo and the other demographic-aware variant (paired $t$-test, $p<0.05$).
        \textit{Avg. Individual Pred.} averages individual predictions across annotators for each item and reports the best score across variant--setting pairs for each metric.
        \textbf{Incorporating socio-demographic information, particularly through explicit reasoning, consistently improves baseline estimation.}
        }
\resizebox{\textwidth}{!}{%
\begin{tabular}{llcccccccc}
\toprule
\multirow{2}{*}{\textbf{Setting}} & \multirow{2}{*}{\textbf{Variant}} & \multicolumn{4}{c}{\textbf{Baseline Estimation (Ann.)}} & \multicolumn{4}{c}{\textbf{Baseline Estimation (Item)}} \\
\cmidrule(lr){3-6}
\cmidrule(lr){7-10}
 &  & MSE $\downarrow$ & MAE $\downarrow$ & $\rho \uparrow$ & W1 $\downarrow$ & MSE $\downarrow$ & MAE $\downarrow$ & $\rho \uparrow$ & W1 $\downarrow$ \\
\midrule
\multirow{3}{*}{Prompted} & NoDemo & 0.067 & 0.195 & 0.524 & 0.114 & 0.065 & 0.194 & 0.542 & 0.115 \\
 & DemoCond & 0.061 & 0.186 & 0.529 & 0.102* & 0.051* & 0.172* & \textbf{0.603*} & 0.101* \\
 & DemoReason & \textbf{0.059} & \textbf{0.183*} & \textbf{0.538} & \textbf{0.097*} & \textbf{0.050*} & \textbf{0.169*} & 0.601* & \textbf{0.097} \\
\midrule
\multirow{3}{*}{Trained} & NoDemo & 0.050 & 0.168 & 0.505 & 0.098 & 0.045 & 0.160 & 0.566 & 0.094 \\
 & DemoCond & 0.047 & 0.164 & \textbf{0.510} & 0.090 & 0.038* & 0.148* & \textbf{0.604*} & 0.086 \\
 & DemoReason & \textbf{0.047} & \textbf{0.163} & 0.502 & \textbf{0.083**} & \textbf{0.037*} & \textbf{0.146*} & 0.599* & \textbf{0.080*} \\
\midrule
\multicolumn{2}{c}{Avg. Individual Pred.} & 0.136 & 0.299 & 0.417 & 0.242 & 0.075 & 0.210 & 0.546 & 0.139 \\
\bottomrule
\end{tabular}
}
\label{tab:crowd}
\end{table}

Table~\ref{tab:crowd} shows that incorporating annotator demographics improves \texttt{crowd} baseline estimation, with the clearest gains at the item level.
In the prompted setting, both DemoCond and DemoReason improve all annotation- and item-level metrics over NoDemo, with almost all item-level gains reaching significance; the trained setting shows largely the same pattern.
The emitted \texttt{crowd} estimates also substantially outperform \textit{Avg. Individual Pred.}, despite having no access to the empirical target distribution, suggesting an advantage of explicitly reasoning of the baseline and modeling the baseline directly as a continuous output rather than recovering it by averaging binary predictions.

Between the two demographic-aware variants, DemoReason matches or improves on DemoCond in MSE, MAE, and W1 at both levels and in both settings.
Its advantage is most evident at the annotation level: DemoReason significantly improves prompted MAE over NoDemo where DemoCond does not, and significantly outperforms both NoDemo and DemoCond on trained W1.
Although DemoReason's improvement in prompted item-level W1 does not reach significance while DemoCond's does, its average W1 remains better.
Explicit demographic reasoning thus strengthens baseline estimation beyond simple conditioning.
Per-model results are provided in Appendix~\ref{app:full_baseline}.

\subsection{Why Does Socio-Demographic Information Improve Baseline Estimation?}
\label{sec:baseline_mech}

Since both DemoCond and DemoReason improve baseline estimation even for prompted LLMs, these gains cannot be explained solely by exposure to empirical baseline distributions during training.
Instead, they suggest that socio-demographic information changes how models reason about and estimate the baseline.
To understand this effect, and the additional role of explicit reasoning, we analyze model behavior and identify two contributing mechanisms.

\paragraph{Incorporating Socio-Demographics Captures Heterogeneity and Reduces Over-Consensus.}
Conditioning on a specific demographic profile leads to less extreme \texttt{crowd} estimates, reducing the model's tendency to assume excessive agreement across annotators.
NoDemo, DemoCond, and DemoReason predict an item-level \texttt{crowd} below 0.1 for 37.1\%, 26.9\%, and 25.6\% of items, respectively, compared with only 16.8\% in the empirical data.
DemoReason reduces this over-consensus most, suggesting that explicitly reasoning about text--annotator interactions makes the model more sensitive to heterogeneous perceptions of the same content.
Consistent with this explanation, Table~\ref{tab:disagreement-item} shows that demographic-aware modeling yields substantially larger improvements on high-disagreement items, suggesting that the baseline gains arise from better modeling annotator heterogeneity.
DemoReason further strengthens these gains over DemoCond.

\begin{table*}[h]
    \centering
    \caption{Gain over NoDemo in item-level baseline estimation, stratified by disagreement
    (low/high disagreement: empirical baseline outside/inside $[0.4,0.6]$, 83.1\%/16.9\% of items). Positive values indicate improvement over NoDemo, with larger values indicating greater gains. \textbf{Demographic-aware variants show larger gains on high-disagreement items.}}
    \label{tab:disagreement-item}
    \resizebox{\textwidth}{!}{%
    \begin{tabular}{llcccccccc}
        \toprule
        \multirow{2}{*}{\textbf{Regime}}
        & \multirow{2}{*}{\textbf{Variant}}
        & \multicolumn{4}{c}{\textbf{Low disagreement}}
        & \multicolumn{4}{c}{\textbf{High disagreement}} \\
        \cmidrule(lr){3-6}
        \cmidrule(lr){7-10}
        & & MSE & MAE & $\rho$ & W1
          & MSE & MAE & $\rho$ & W1 \\
        \midrule

        \multirow{2}{*}{Prompted}
            & DemoCond
            & +0.013 & +0.019 & \textbf{+0.075} & +0.013
            & +0.020 & +0.041 & \textbf{+0.022} & +0.042 \\

            & DemoReason
            & \textbf{+0.014} & \textbf{+0.019} & +0.069 & \textbf{+0.014}
            & \textbf{+0.026} & \textbf{+0.052} & +0.017 & \textbf{+0.052} \\

        \midrule

        \multirow{2}{*}{Trained}
            & DemoCond
            & +0.003 & +0.006 & +0.034 & +0.009
            & +0.012 & +0.025 & +0.024 & \textbf{+0.024} \\

            & DemoReason
            & \textbf{+0.005} & \textbf{+0.008} & \textbf{+0.048} & \textbf{+0.013}
            & \textbf{+0.014} & \textbf{+0.031} & \textbf{+0.091} & +0.023 \\

        \bottomrule
    \end{tabular}%
    }
\end{table*}

\begin{wrapfigure}{r}{0.48\linewidth}
\vspace{-4mm}
    \centering
        \includegraphics[width=\linewidth]{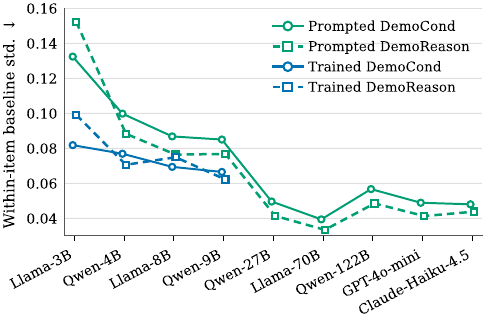}
        \caption{Within-item standard deviation of baseline estimates for DemoCond and DemoReason across trained and prompted LLMs. \textbf{DemoReason shows lower within-item variation than DemoCond for prompted models.}}
        \label{fig:crowd variance}
\vspace{-4mm}
\end{wrapfigure}

\paragraph{Reasoning Makes Baseline Estimates More Stable Across Annotators.}
A second mechanism helps explain why Demo\-Reason outperforms DemoCond on annotation-level baseline metrics, particularly for prompted LLMs.
By reasoning about demographic-specific perceptions in a separate block, DemoReason better separates the annotator-specific component from the \texttt{crowd} baseline, which should be invariant to the target annotator.
It therefore produces more consistent \texttt{crowd} estimates across annotators of the same item, as reflected by the lower within-item standard deviation in Figure~\ref{fig:crowd variance}.
The greater stability allows item-level improvements to carry over to the annotation level.
This advantage is not observed for trained models, whose within-item \texttt{crowd} variation is already lower than that of their prompted counterparts.

\vspace{-0.5em}
\begin{center}
\colorbox{gray!15}{%
\parbox{0.98\linewidth}{%
\textbf{Takeaway}: Incorporating annotator socio-demographics improves baseline estimation by making models more sensitive to heterogeneous perceptions, while explicit reasoning further improves estimation by making the inferred baseline more stable across annotators.}}
\end{center}
\vspace{-0.5em}

\subsection{LLMs Capture Meaningful Demographic-Specific Deviations}
\label{sec:shift_eval}
We next ask whether the model-emitted \texttt{shift} captures systematic demographic-specific deviations from the baseline.
Since neither the true deviation $\delta^\star(x_i, \mathbf z_i^j)$ nor its empirical counterpart $\delta(x_i, \mathbf z_i^j)$ is directly observable, we evaluate \texttt{shift} against three increasingly fine-grained proxy targets: global group tendencies, individual--majority deviations, and item--group interactions.

\paragraph{Reasoning-Enhanced Shift Predictions Recover Global Group Tendencies}

We first test whether \texttt{shift} predictions recover dataset-level group tendencies.
For each annotation of item $i$ by annotator $j$, we define a noisy individual-level shift proxy as $y_i^j-c(x_i)$, i.e., the deviation of the individual annotation from the empirical baseline.
Averaging these shifts over annotations from demographic group $g$ yields its empirical global tendency $\Delta_g$, indicating whether members of $g$ tend to judge their assigned items as more or less offensive than the corresponding baselines.
We analogously obtain $\hat{\Delta}_g$ by averaging the model-emitted \texttt{shift} for DemoCond and DemoReason; for NoDemo, which does not explicitly predict a \texttt{shift}, we instead use $\hat{y}_i^j-\hat{c}(x_i)$.
We then compare $\hat{\Delta}_g$ with $\Delta_g$ using Pearson correlation (\textbf{$r$}), Spearman correlation (\textbf{$\rho$}), and balanced sign agreement (\textbf{Sign}) across groups within each demographic axis.

\begin{figure}
    \centering
    \includegraphics[width=\linewidth]{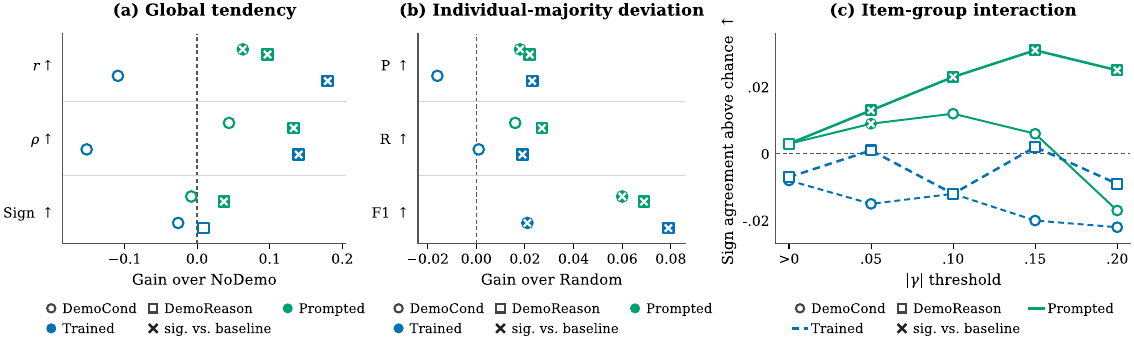}
    \caption{Model-emitted \texttt{shift} results for (a) global group tendency, shown as gain over NoDemo; (b) individual--majority deviations, shown as gain over random; and (c) item--group interactions, shown as sign agreement above chance across $|\gamma|$ thresholds. Scores are averaged over dataset--model pairs; significance is assessed with paired $t$-tests ($p<0.05$).
   \textbf{DemoReason consistently captures meaningful demographic effects across all three analyses.}}
    \label{fig:shift_analysis}
    \vspace{-5mm}
\end{figure}

Figure~\ref{fig:shift_analysis}(a) shows that DemoReason recovers dataset-level group tendencies: it outperforms both NoDemo and DemoCond across all metrics and settings, with significant gains in all cases except trained sign agreement.
In contrast, DemoCond yields only limited and inconsistent gains for prompted models and even underperforms NoDemo after training.
One possible explanation is that DemoCond predicts \texttt{shift}=0 for the vast majority of instances, suggesting that demographic conditioning alone often fails to elicit demographic-specific deviations.
These results suggest that LLMs can leverage socio-demographic information to recover systematic group tendencies, but that doing so relies on explicit reasoning rather than conditioning alone.

\paragraph{Reasoning-Enhanced Shifts Capture Individual--Majority Deviations Beyond Chance.}

We next examine \texttt{shift} at a finer granularity using a three-way proxy target based on individual--majority prediction deviation to approximate the underlying demographic-specific deviation.
The proxy is $0$ when the individual annotation agrees with the empirical majority, $+1$ when the majority is non-offensive but the individual annotation is offensive, and $-1$ in the opposite direction.\footnote{This is a noisy proxy for the demographic-specific deviation $\delta^\star(x_i, \mathbf z_i^j)$, as it inherits the individual residual $\varepsilon_i^j$ and captures only effects large enough to change the final discrete prediction.}
Performance above a random baseline would indicate that \texttt{shift} captures item-specific demographic signal informative of individual--majority deviations. Figure~\ref{fig:shift_analysis}(b) reports macro-averaged precision, recall, and F1 over the three proxy classes.

In line with the global tendency results, DemoReason consistently outperforms the random baseline, whereas DemoCond shows no consistent improvement, particularly after training.
The gains are nevertheless modest, consistent with Section~\ref{sec:residual dominance}, where socio-demographic attributes explain only a small fraction of individual-level variation.

\paragraph{Reasoning Identifies Item--Group Interactions Beyond Global Group Tendency}

An observed demographic-specific deviation may reflect both a group's global tendency and an item-specific \textit{interaction} with the text.
We therefore ask whether \texttt{shift} captures these finer-grained interactions beyond global tendencies.
For example, although annotators from Japan are globally less likely to rate texts as offensive, DemoReason predicts a positive shift for the example in Figure~\ref{fig:reasoning_example}, capturing how the specific content interacts with the annotator's demographics.
Since such interactions are sparse and lack explicit ground-truth labels, we focus on item--group pairs whose demographic effects deviate substantially from the corresponding global tendency.

To quantify such interactions, for each item $i$ and group $g$ with at least $m=5$ annotations, we first define the observed deviation as $\delta_{i,g}=c_g(x_i)-c(x_i)$, where $c_g(x_i)$ and $c(x_i)$ are the group-specific and overall empirical baseline, respectively.
We then subtract the group's global tendency $\Delta_g$ to obtain the item-specific interaction $\gamma_{i,g}=\delta_{i,g}-\Delta_g$.
Analogously, from the model-emitted \texttt{shift}, we compute $\hat{\gamma}_{i,g}=\bar{\texttt{shift}}_{i,g}-{\hat{\Delta}}_g$, where $\bar{\texttt{shift}}_{i,g}$ is the mean predicted shift for the item--group pair. 
We evaluate balanced sign agreement between $\gamma_{i,g}$ and $\hat{\gamma}_{i,g}$ at increasing $|\gamma|$ thresholds, progressively focusing on pairs with stronger item--group interactions.

Figure~\ref{fig:shift_analysis}(c) shows that prompted DemoReason is the only variant to perform significantly above chance on item--group interactions.
Its performance further improves as the $|\gamma|$ threshold increases, consistent with stronger interactions being less affected by annotation noise.
However, its improvement over chance remains below 0.04, indicating that the captured interaction signal is modest.
These results suggest that, with explicit reasoning, prompted LLMs can identify fine-grained interactions between text content and annotator socio-demographics.

\vspace{-0.5em}
\begin{center}
\colorbox{gray!15}{%
\parbox{0.98\linewidth}{%
\textbf{Takeaway}: With explicit reasoning about socio-demographic information, LLMs capture meaningful demographic-specific deviations that do not consistently emerge under simple conditioning.}}
\end{center}
\vspace{-0.5em}

\subsection{Better Baseline and Deviation Estimates Improve Final Predictions}
\label{sec:intervention}

Finally, we examine whether better baseline and deviation estimates translate into better final predictions, using observational and interventional evidence on POPQUORN.

\paragraph{Observational Evidence}
We first provide observational evidence by correlating prediction correctness with the quality of \texttt{crowd} and \texttt{shift} estimates.
We measure \texttt{crowd} quality as the negative squared error from the empirical baseline, and \texttt{shift} correctness against the three-way individual--majority deviation proxy.
Across trained and prompted DemoCond and DemoReason models, individual prediction correctness is positively correlated with both \texttt{crowd} quality ($r=0.24$--$0.27$) and \texttt{shift} correctness ($r=0.40$--$0.57$).

\paragraph{Interventional Evidence}
We intervene on the \texttt{crowd} and \texttt{shift} estimates of a trained and a prompted LLM (Llama3.1-8B and Llama3.3-70B) and measure the resulting changes in final predictions.
An \textbf{oracle} intervention replaces \texttt{crowd} with the empirical baseline and \texttt{shift} with the individual--majority deviation proxy, whereas an \textbf{adversarial} intervention replaces both with estimates from a randomly sampled item.
As shown in Table~\ref{tab:intervention-joint}, oracle interventions mostly improve prediction performance, while adversarial interventions generally degrade it.
Component-wise interventions (Appendix~\ref{app:intervention_test}) show a clearer effect for \texttt{crowd}, whereas the original \texttt{shift} predictions do not consistently outperform adversarial shifts, consistent with the modest shift effects in Section~\ref{sec:shift_eval}.
Together, these results indicate that more accurate baseline and shift estimates are associated with better predictions, and that the quality of these components directly influences final predictions.

\begin{table*}[h]
\centering
\small
\caption{Performance under joint interventions on \texttt{crowd} and \texttt{shift} estimates. 
Each cell reports \textit{Original / Oracle / Adversarial} performance.
Bold marks cells following the expected ordering Oracle $>$ Original $>$ Adversarial.
\textbf{Better baseline and shift estimates are not only associated with, but also lead to, improved final predictions.}}
\label{tab:intervention-joint}
\resizebox{\textwidth}{!}{%
\begin{tabular}{llccccc}
\toprule
\multirow{2}{*}{\textbf{Model}} & \multirow{2}{*}{\textbf{Variant}} & \textbf{Individual} & \multicolumn{2}{c}{\textbf{Group}} & \multicolumn{2}{c}{\textbf{Population}} \\
\cmidrule(lr){3-3}
\cmidrule(lr){4-5}
\cmidrule(lr){6-7}
 & & F1 $\uparrow$ & F1 $\uparrow$ & MSE $\downarrow$ & F1 $\uparrow$ & MSE $\downarrow$ \\
\midrule
        \multirow{2}{*}{Llama-8B (Trained)} & DemoCond & \textbf{0.453/0.540/0.364} & \textbf{0.442/0.589/0.424} & 0.190/0.105/0.170 & \textbf{0.549/0.667/0.444} & \textbf{0.094/0.084/0.106} \\
         & DemoReason & \textbf{0.474/0.600/0.470} & \textbf{0.450/0.522/0.439} & \textbf{0.188/0.145/0.196} & 0.526/0.600/0.541 & 0.059/0.065/0.068 \\
         \midrule
        \multirow{2}{*}{Llama-70B (Prompted)} & DemoCond & \textbf{0.438/0.554/0.312} & \textbf{0.425/0.544/0.257} & \textbf{0.178/0.140/0.200} & \textbf{0.435/0.652/0.387} & 0.095/0.068/0.077 \\
         & DemoReason & \textbf{0.458/0.557/0.401} & \textbf{0.443/0.550/0.379} & \textbf{0.180/0.166/0.201} & \textbf{0.500/0.630/0.465} & \textbf{0.084/0.068/0.085} \\
\bottomrule
\end{tabular}%
}
\end{table*}

\section{Conclusion}

We analyzed when and where socio-demographic information benefits pluralistic offensiveness modeling, together with the role of explicit reasoning.
The value of socio-demographics lies more in capturing systematic group- and population-level patterns than in recovering individual judgments, whose idiosyncratic variation can obscure these gains until it is reduced through aggregation.
These benefits arise from both improved baseline estimation and meaningful, though modest, demographic-specific deviation estimates, while explicit reasoning about socio-demographic information strengthens both components beyond direct conditioning and leads to better predictions.
Limited individual-level gains should thus not be interpreted as evidence that socio-demographic information is uninformative.
We encourage future work to evaluate socio-demographic modeling across aggregation levels and, where possible, across distinct prediction components.
\footnote{Limitations of our work are discussed in Appendix~\ref{app:limitations}.}

\section*{Acknowledgements}
This work was funded by the DFG project GRK 2853 "Neuroexplicit Models of Language, Vision, and Action" (project number 471607914).

\section*{AI use statement}

In this work, we used generative AI tools to assist with code development, related work discovery, and the editing and polishing of the manuscript.
We did not use generative AI tools to formulate the research ideas or interpret the experimental results; these were developed and analyzed by the authors.
All AI-assisted code and text were reviewed and verified by the authors.
We take full responsibility for the final content of this work.

\section*{Reproducibility statement}

We provide the details necessary to reproduce our experiments in the main paper, Appendix, and supplementary materials, including model configurations, prompting and training procedures, evaluation protocols, and implementation details. 
The complete codebase will be publicly released upon acceptance of the paper.

\section*{Ethics statement}

Using socio-demographic information in predictive models warrants careful consideration. 
Demographic attributes may introduce broad group-level associations or stereotypes that do not always reflect individual annotators accurately.
Such information should therefore be used cautiously and should not be interpreted as deterministically implying an individual's beliefs or judgments.
Our explicit reasoning format makes the use of demographic information more transparent by exposing the model's baseline and demographic-specific reasoning, which can facilitate inspection and auditing.
However, transparency does not guarantee that the reasoning is unbiased or appropriate, and the generated rationales themselves may reproduce stereotypes.
We therefore recommend careful evaluation and human inspection when demographic-aware models are deployed or used to support real-world decisions.

\bibliography{iclr2027_conference}
\bibliographystyle{iclr2027_conference}

\appendix

\section{Related Work}

\paragraph{Socio-Demographics in Pluralistic Modeling}
\label{app:related work}
Human values are inherently heterogeneous~\citep{maxmin-rlhf-2024, shen-etal-2025-micro, wang2026sparse, wang2026routingweightsfaithfulresponselevel}, giving rise to diverse and subjective perceptions in pluralistic tasks such as bias, hate speech, and offensiveness detection~\citep{10.1145/3491102.3502004, fleisig-etal-2023-majority, wang-demberg-2024-parameter, wang-demberg-2024-rsa, wang2025bcos,  wang_bridging_fairness_explainability}.
Prior work has explored incorporating annotators' socio-demographic information into pluralistic modeling at multiple granularities, with mixed results.
Some approaches report gains for certain demographic axes or experimental settings~\citep{xu-etal-2026-modeling, zhang-coltekin-2026-modeling}, whereas others find limited benefits or even negative effects~\citep{orlikowski-etal-2023-ecological, orlikowski-etal-2025-beyond, sorensen-etal-2025-value}.
These findings are consistent with our observation that demographic information does not reliably improve modeling at lower levels of aggregation.
\citet{cai2026does} further investigates the data and training conditions under which demographic modeling benefits individual-level prediction.
However, existing work largely treats individual, group, and population modeling as separate evaluation targets, rather than systematically characterizing when the benefits of demographic information become more visible across levels of aggregation.
Moreover, prior work typically evaluates only the resulting predictions, leaving unclear where the benefits of socio-demographic information arise within the prediction process: whether they come from better estimation of a shared overall baseline, demographic-specific shifts, or both.
Consequently, it provides limited insight into how demographic information contributes to improved pluralistic modeling.

\paragraph{Reasoning in LLMs}

LLMs with reasoning capabilities have achieved substantial gains in tasks such as mathematics and coding~\citep{jaech2024openai, guo2025deepseek}.
More recently, reasoning has also been introduced into subjective tasks, including reward modeling and bias detection~\citep{ICLR2026_8e3b8de2, fan-etal-2025-biasguard}.
These approaches typically first distill reasoning traces from stronger reasoning models to establish the desired format, and then further optimize the model with RL objectives such as DPO or GRPO.
However, existing work generally evaluates reasoning as a complete modeling paradigm, without isolating whether reasoning about a particular source of information changes how that information is used by the model.
Our work instead examines how explicit reasoning about socio-demographics affects distinct components of pluralistic modeling, including baseline estimation, demographic-specific shift estimation, and the final prediction.

\section{Details on Experimental Setup}
\label{app:details}

\begin{table}[t]
\centering
\scriptsize
\caption{Dataset statistics and socio-demographic coverage. 
}
\begin{tabular}{lccc}
\toprule
\textbf{Dataset} & \textbf{Test Annots. / Items} & \textbf{ \# Annots. / Item} & \textbf{Axes (\# groups)} \\
\midrule
D3CODE  & 2,507 / 77 & 32.6 & region (8), country (21), age (3), gender (3), SES (10) \\
DICES  & 10,849 / 134 & 81.0 & gender (4), race (6), age (3), education (3) \\
POPQUORN  & 1,308 / 150 & 8.7 & gender (3), race (6), age (10), education (5) \\
\bottomrule
\end{tabular}

\label{tab:datasets}
\end{table}

\paragraph{Datasets}

The dataset statistics and socio-demographic coverage of the three multi-annotator datasets are summarized in Table~\ref{tab:datasets}. 
We report only test-set statistics, since the effective amount of training data varies with the number of SFT and GRPO optimization steps.

\paragraph{Models}

We use both open-source and commercial models for training and prompting. 
For training, we consider Llama3.2-3B, Llama3.1-8B, Qwen3.5-4B, and Qwen3.5-9B. 
For prompting, we additionally evaluate Qwen3.5-27B, Llama3.3-70B, Qwen3.5-122B-A10B, GPT-4o-mini, and Haiku-4.5. We use the instruction-tuned variants of all Llama models. 
For Qwen models, thinking mode is disabled during training. 
In the prompting setting, we likewise disable thinking mode for all models except Qwen3.5-122B-A10B, as enabling it substantially reduces output parseability for the smaller models.

\paragraph{DemoReason Prompt}

We present the complete DemoReason prompt below. DemoCond and NoDemo are
constructed by removing specific components from this full prompt, as described
afterward.

\textbf{System prompt.}

\begin{minipage}{\linewidth}
\small\ttfamily\raggedright
You are a careful fairness and toxicity judge. Given a social media post preceded
by an [audience: ...] annotator profile, produce structured reasoning in tagged
blocks --- <read>, <norm>, <baseline>, <audience> --- and end with a single line
in exactly this format:

=== Verdict: target=<group|none>, crowd=<0.0-1.0>,
shift=<-1|0|+1>, offensive=<yes|no>

shift encodes DIRECTION ONLY: +1 if this annotator's group finds the post MORE
offensive than the broad pool, -1 if LESS, 0 if no meaningful difference.
\end{minipage}

\vspace{0.5em}

\textbf{User turn.}

\begin{minipage}{\linewidth}
\small\ttfamily\raggedright
You are annotating social media posts for offensiveness on behalf of a specific
annotator. Posts are rated by many annotators, who do not always agree. Decide
how this post would be rated by the annotator described below.

POST: \{text\}

ANNOTATOR: [audience: \{axis\_1\}=\{val\_1\},
\{axis\_2\}=\{val\_2\}, ...]

<read> what is said, literally and by implication, and about whom

<norm> what norm would make this offensive, if any, and whose norm

<baseline> how a broad, diverse pool of annotators would split on this, and why

<audience> whether this annotator's group tends to read THIS post differently
from the broad pool, and why. shift encodes DIRECTION ONLY and must be exactly
one of three values: +1 if the group would find it MORE offensive than the
broad pool, -1 if LESS, 0 if no meaningful difference. It is a direction, not
an intensity: never write +2, -2, or any other number, no matter how large the
difference feels. 0 is the correct, expected answer whenever the group reads
the post about the same as the pool.

=== Verdict: target=<jewish|christian|muslim|LGB|transgender|none>,
crowd=<0.0-1.0>, shift=<-1|0|+1>, offensive=<yes|no>

Rules:

- Reason as genuine analysis.

- Never infer an opinion from an identity ("as a woman, she would...").

  Reason about norms, context, and what registers as ordinary speech.
  
- Report crowd as the fraction of a broad, diverse pool that would find the
  post offensive, rounded to one decimal place: one of
  0.0, 0.1, 0.2, ..., 1.0.
  
- In the verdict line write bare values, never the angle brackets from the
  specification above.
  
- Use 1--3 sentences per block and use the tags exactly.

- The verdict line must be the final line of the response.
\end{minipage}

\paragraph{DemoCond Prompt}
DemoCond removes the explicit \texttt{<audience>} reasoning block from
DemoReason while retaining the annotator socio-demographic profile and the
\texttt{shift} prediction. Accordingly, the required reasoning blocks are
\texttt{<read>}, \texttt{<norm>}, and \texttt{<baseline>}; all other task
instructions, demographic inputs, output fields, and rules remain unchanged.

\paragraph{NoDemo Prompt}
NoDemo further removes all annotator-specific information. The
\texttt{ANNOTATOR} profile is omitted, the task is framed without reference to
a specific annotator, and both the \texttt{<audience>} block and
\texttt{shift} field are removed. The remaining reasoning blocks are
\texttt{<read>}, \texttt{<norm>}, and \texttt{<baseline>}, followed by
\texttt{target}, \texttt{crowd}, and \texttt{offensive}.

\paragraph{Training Configuration}

We use identical training hyperparameters across all datasets and variants. 
For SFT, we apply QLoRA with 4-bit quantization (LoRA rank 16, $\alpha=32$), a learning rate of $2\times10^{-4}$, 3 epochs, an effective batch size of 32, and a maximum sequence length of 1024.
GRPO is initialized from the corresponding SFT checkpoint and trained for 150 steps with a learning rate of $1\times10^{-6}$, batch size 8 with gradient accumulation 16, 8 generations per prompt with temperature=$1.0$, $\beta=0.001$, and a maximum completion length of 768. 
Importantly, GRPO uses only final-label correctness as the reward, based on the emitted \texttt{offensive} prediction, without any explicit reward on \texttt{crowd} or \texttt{shift}.
At inference time, we use greedy decoding.

\paragraph{Baselines}

We compare against several baselines to contextualize our results.
For prompted LLMs, we consider zero-shot and free-form CoT baselines, both with and without user profiles in the input. 
All prompted baselines are required to output a \texttt{crowd} field as their estimate of the baseline.
For training, we compare BERT models that are annotator-blind with majority-vote or individual labels~\citep{davani-etal-2022-dealing}, and annotator-aware with textual demographic descriptions~\citep{fleisig-etal-2023-majority} or learned demographic embeddings~\citep{10.1145/3491102.3502004}. 
We also include DEM-MoE~\citep{xu-etal-2026-modeling}, a more recent method that uses external LLM judgments during training.
The predicted positive-class probability is used as the baseline estimate.

\section{Full Performance Results}
\label{app:full_performance}

Tables~\ref{tab:main-prompted} and~\ref{tab:main-trained} report per-model performance at the individual, group, and population levels.
The baseline models exhibit the same overall pattern: the benefits of incorporating socio-demographics are limited at the individual level but become increasingly apparent with aggregation.

\begin{table}[t]
\centering
\small
\caption{
        Performance for all \textit{prompted} models and baselines across different levels.
        Metric scores are averaged over three datasets, and \textit{Overall} averages across backbones.
        * denotes significant improvement over NoDemo; ** denotes significant improvement over both NoDemo and the other variant across model--dataset pairs (paired $t$-test, $p<0.05$).
        }
\resizebox{\textwidth}{!}{%
\begin{tabular}{llcccccccccc}
\toprule
\multirow{2}{*}{\textbf{Model}} & \multirow{2}{*}{\textbf{Variant}} & \multicolumn{2}{c}{\textbf{Individual}} & \multicolumn{4}{c}{\textbf{Group}} & \multicolumn{4}{c}{\textbf{Population}} \\
\cmidrule(lr){3-4}
\cmidrule(lr){5-8}
\cmidrule(lr){9-12}
 &  & F1 $\uparrow$ & Acc $\uparrow$ & F1$\uparrow$ & Acc$\uparrow$ & MSE$\downarrow$ & $\rho\uparrow$ & F1 $\uparrow$ & Acc $\uparrow$ & MSE $\downarrow$ & $\rho \uparrow$ \\
\midrule
\multirow{3}{*}{Llama-3B} & NoDemo & \textbf{0.455} & 0.583 & 0.376 & 0.623 & 0.377 & 0.247 & 0.324 & 0.645 & 0.302 & 0.401 \\
 & DemoCond & 0.453 & \textbf{0.584} & \textbf{0.380} & \textbf{0.624} & 0.320 & \textbf{0.271} & 0.334 & 0.639 & 0.194 & \textbf{0.533} \\
 & DemoReason & 0.425 & 0.569 & 0.360 & 0.608 & \textbf{0.280} & 0.227 & \textbf{0.337} & \textbf{0.663} & \textbf{0.134} & 0.498 \\
\midrule
\multirow{3}{*}{Qwen-4B} & NoDemo & \textbf{0.426} & \textbf{0.593} & \textbf{0.391} & \textbf{0.638} & 0.260 & \textbf{0.282} & \textbf{0.384} & \textbf{0.682} & 0.189 & 0.443 \\
 & DemoCond & 0.364 & 0.575 & 0.362 & 0.618 & 0.211 & 0.270 & 0.360 & 0.664 & 0.115 & 0.494 \\
 & DemoReason & 0.346 & 0.567 & 0.342 & 0.609 & \textbf{0.200} & 0.262 & 0.335 & 0.648 & \textbf{0.095} & \textbf{0.542} \\
\midrule
\multirow{3}{*}{Llama-8B} & NoDemo & 0.406 & \textbf{0.570} & 0.352 & 0.605 & 0.271 & 0.210 & \textbf{0.377} & 0.659 & 0.197 & 0.356 \\
 & DemoCond & 0.397 & 0.561 & 0.348 & 0.602 & \textbf{0.221} & 0.223 & 0.361 & 0.659 & \textbf{0.089} & 0.490 \\
 & DemoReason & \textbf{0.474} & 0.553 & \textbf{0.371} & \textbf{0.609} & 0.342 & \textbf{0.277} & 0.343 & \textbf{0.665} & 0.202 & \textbf{0.564} \\
\midrule
\multirow{3}{*}{Qwen-9B} & NoDemo & 0.358 & 0.572 & 0.361 & 0.618 & 0.229 & 0.229 & 0.356 & 0.651 & 0.162 & 0.381 \\
 & DemoCond & 0.436 & 0.602 & \textbf{0.420} & \textbf{0.663} & \textbf{0.227} & 0.314 & \textbf{0.424} & 0.714 & 0.133 & 0.548 \\
 & DemoReason & \textbf{0.456} & \textbf{0.610} & 0.418 & 0.662 & 0.232 & \textbf{0.326} & 0.414 & \textbf{0.723} & \textbf{0.128} & \textbf{0.568} \\
\midrule
\multirow{3}{*}{Qwen-27B} & NoDemo & 0.441 & 0.609 & 0.422 & 0.660 & 0.228 & 0.317 & 0.428 & 0.718 & 0.158 & 0.507 \\
 & DemoCond & 0.449 & 0.611 & 0.437 & 0.668 & \textbf{0.219} & 0.338 & 0.453 & 0.730 & 0.132 & 0.553 \\
 & DemoReason & \textbf{0.457} & \textbf{0.614} & \textbf{0.442} & \textbf{0.678} & 0.222 & \textbf{0.347} & \textbf{0.454} & \textbf{0.734} & \textbf{0.130} & \textbf{0.576} \\
\midrule
\multirow{3}{*}{Llama-70B} & NoDemo & 0.411 & 0.596 & 0.412 & 0.655 & 0.219 & 0.296 & 0.427 & 0.701 & 0.150 & 0.471 \\
 & DemoCond & 0.417 & 0.599 & 0.420 & 0.657 & 0.201 & \textbf{0.331} & 0.413 & 0.690 & 0.127 & 0.526 \\
 & DemoReason & \textbf{0.420} & \textbf{0.600} & \textbf{0.421} & \textbf{0.660} & \textbf{0.199} & 0.330 & \textbf{0.432} & \textbf{0.704} & \textbf{0.118} & \textbf{0.533} \\
\midrule
\multirow{3}{*}{Qwen-122B} & NoDemo & 0.358 & 0.579 & 0.389 & 0.637 & 0.195 & 0.291 & 0.420 & 0.683 & 0.134 & 0.433 \\
 & DemoCond & 0.377 & 0.582 & 0.386 & 0.628 & 0.191 & 0.312 & 0.392 & 0.659 & \textbf{0.098} & 0.510 \\
 & DemoReason & \textbf{0.400} & \textbf{0.589} & \textbf{0.407} & \textbf{0.646} & \textbf{0.189} & \textbf{0.325} & \textbf{0.438} & \textbf{0.686} & 0.100 & \textbf{0.518} \\
\midrule
\multirow{3}{*}{GPT-4o-mini} & NoDemo & \textbf{0.476} & 0.601 & 0.372 & 0.612 & 0.375 & 0.281 & 0.329 & 0.647 & 0.291 & 0.495 \\
 & DemoCond & 0.466 & \textbf{0.607} & 0.390 & 0.638 & 0.285 & \textbf{0.312} & 0.383 & 0.700 & 0.181 & \textbf{0.583} \\
 & DemoReason & 0.460 & 0.600 & \textbf{0.394} & \textbf{0.640} & \textbf{0.257} & 0.298 & \textbf{0.405} & \textbf{0.716} & \textbf{0.135} & 0.542 \\
\midrule
\multirow{3}{*}{Haiku-4.5} & NoDemo & 0.356 & 0.579 & 0.366 & 0.621 & 0.209 & 0.269 & \textbf{0.415} & 0.670 & 0.141 & 0.421 \\
 & DemoCond & 0.368 & 0.579 & 0.368 & 0.625 & 0.200 & 0.282 & 0.383 & 0.664 & 0.116 & 0.500 \\
 & DemoReason & \textbf{0.398} & \textbf{0.588} & \textbf{0.390} & \textbf{0.637} & \textbf{0.198} & \textbf{0.317} & 0.388 & \textbf{0.672} & \textbf{0.111} & \textbf{0.544} \\
\midrule
\multirow{3}{*}{Overall} & NoDemo & 0.410 & 0.587 & 0.382 & 0.630 & 0.263 & 0.269 & 0.384 & 0.673 & 0.191 & 0.434 \\
 & DemoCond & 0.414 & \textbf{0.589} & 0.390 & 0.636 & \textbf{0.230*} & 0.295* & 0.389 & 0.680* & 0.132* & 0.526* \\
 & DemoReason & \textbf{0.426**} & 0.588 & \textbf{0.394*} & \textbf{0.639} & 0.235* & \textbf{0.301*} & \textbf{0.394*} & \textbf{0.690**} & \textbf{0.128*} & \textbf{0.543**} \\
\midrule
\multicolumn{12}{c}{\textit{Baselines}} \\
\midrule
\multicolumn{12}{l}{\textbf{GPT-4o-mini}} \\
\multicolumn{2}{l}{Zero-shot} & \textbf{0.450} & 0.586 & \textbf{0.411} & \textbf{0.650} & 0.265 & 0.296 & \textbf{0.389} & \textbf{0.690} & 0.188 & 0.485 \\
\multicolumn{2}{l}{Zero-shot + Demo.} & 0.436 & \textbf{0.598} & 0.405 & 0.644 & 0.211 & \textbf{0.333} & 0.349 & 0.643 & 0.108 & \textbf{0.586} \\
\multicolumn{2}{l}{CoT} & 0.422 & 0.568 & 0.345 & 0.590 & 0.312 & 0.202 & 0.335 & 0.656 & 0.233 & 0.327 \\
\multicolumn{2}{l}{CoT + Demo.} & 0.432 & 0.577 & 0.401 & 0.642 & \textbf{0.209} & 0.324 & 0.369 & 0.654 & \textbf{0.083} & 0.573 \\
\midrule
\multicolumn{12}{l}{\textbf{Haiku-4.5}} \\
\multicolumn{2}{l}{Zero-shot} & 0.300 & 0.563 & 0.345 & 0.615 & 0.189 & 0.260 & 0.343 & 0.619 & 0.130 & 0.372 \\
\multicolumn{2}{l}{Zero-shot + Demo.} & 0.242 & 0.546 & 0.296 & 0.593 & \textbf{0.175} & 0.261 & 0.373 & 0.629 & 0.095 & 0.423 \\
\multicolumn{2}{l}{CoT} & \textbf{0.395} & \textbf{0.582} & \textbf{0.379} & \textbf{0.627} & 0.232 & 0.257 & \textbf{0.382} & \textbf{0.661} & 0.164 & 0.408 \\
\multicolumn{2}{l}{CoT + Demo.} & 0.378 & 0.577 & 0.364 & 0.621 & 0.185 & \textbf{0.305} & 0.366 & 0.640 & \textbf{0.079} & \textbf{0.533} \\
\bottomrule
\end{tabular}
}
\label{tab:main-prompted}
\end{table}

\begin{table}[h]
\centering
\small
\caption{
        Performance for all \textit{trained} models and baselines across different levels.
        Metric scores are averaged over three datasets, and \textit{Overall} averages across backbones.
        * denotes significant improvement over NoDemo; ** denotes significant improvement over both NoDemo and the other variant across model--dataset pairs (paired $t$-test, $p<0.05$).
        }
\resizebox{\textwidth}{!}{%
\begin{tabular}{llcccccccccc}
\toprule
\multirow{2}{*}{\textbf{Model}} & \multirow{2}{*}{\textbf{Variant}} & \multicolumn{2}{c}{\textbf{Individual}} & \multicolumn{4}{c}{\textbf{Group}} & \multicolumn{4}{c}{\textbf{Population}} \\
\cmidrule(lr){3-4}
\cmidrule(lr){5-8}
\cmidrule(lr){9-12}
 &  & F1 $\uparrow$ & Acc $\uparrow$ & F1$\uparrow$ & Acc$\uparrow$ & MSE$\downarrow$ & $\rho\uparrow$ & F1 $\uparrow$ & Acc $\uparrow$ & MSE $\downarrow$ & $\rho \uparrow$ \\
\midrule
\multirow{3}{*}{Llama-3B} & NoDemo & \textbf{0.342} & \textbf{0.576} & 0.339 & 0.610 & 0.204 & 0.261 & 0.340 & 0.634 & 0.124 & 0.424 \\
 & DemoCond & 0.308 & 0.562 & 0.312 & 0.592 & 0.188 & 0.239 & 0.310 & 0.600 & 0.097 & 0.439 \\
 & DemoReason & 0.334 & 0.570 & \textbf{0.356} & \textbf{0.611} & \textbf{0.188} & \textbf{0.282} & \textbf{0.378} & \textbf{0.657} & \textbf{0.073} & \textbf{0.497} \\
\midrule
\multirow{3}{*}{Qwen-4B} & NoDemo & 0.381 & 0.587 & 0.404 & \textbf{0.645} & 0.197 & 0.317 & 0.467 & \textbf{0.720} & 0.117 & 0.512 \\
 & DemoCond & 0.372 & \textbf{0.589} & 0.402 & 0.642 & \textbf{0.171} & \textbf{0.331} & 0.444 & 0.686 & 0.094 & \textbf{0.540} \\
 & DemoReason & \textbf{0.389} & 0.586 & \textbf{0.405} & 0.645 & 0.185 & 0.314 & \textbf{0.526} & \textbf{0.761} & \textbf{0.079} & 0.537 \\
\midrule
\multirow{3}{*}{Llama-8B} & NoDemo & 0.403 & \textbf{0.599} & 0.395 & 0.648 & 0.215 & 0.304 & 0.411 & 0.712 & 0.137 & 0.495 \\
 & DemoCond & \textbf{0.408} & 0.597 & \textbf{0.411} & \textbf{0.656} & 0.185 & \textbf{0.331} & \textbf{0.439} & \textbf{0.726} & 0.098 & \textbf{0.567} \\
 & DemoReason & 0.365 & 0.534 & 0.384 & 0.635 & \textbf{0.176} & 0.316 & 0.366 & 0.649 & \textbf{0.069} & 0.519 \\
\midrule
\multirow{3}{*}{Qwen-9B} & NoDemo & 0.303 & 0.570 & 0.360 & 0.621 & 0.175 & 0.282 & 0.430 & 0.668 & 0.102 & 0.461 \\
 & DemoCond & \textbf{0.333} & \textbf{0.574} & \textbf{0.374} & \textbf{0.626} & \textbf{0.170} & \textbf{0.309} & \textbf{0.451} & \textbf{0.680} & 0.090 & 0.502 \\
 & DemoReason & 0.315 & 0.569 & 0.353 & 0.621 & 0.176 & 0.279 & 0.416 & 0.679 & \textbf{0.080} & \textbf{0.523} \\
\midrule
\multirow{3}{*}{Overall} & NoDemo & \textbf{0.357} & \textbf{0.583} & 0.375 & \textbf{0.631} & 0.198 & 0.291 & 0.412 & 0.684 & 0.120 & 0.473 \\
 & DemoCond & 0.355 & 0.580 & \textbf{0.375} & 0.629 & \textbf{0.179*} & \textbf{0.303*} & 0.411 & 0.673 & 0.095* & 0.512* \\
 & DemoReason & 0.351 & 0.565 & 0.375 & 0.628 & 0.181* & 0.298* & \textbf{0.421*} & \textbf{0.686} & \textbf{0.075**} & \textbf{0.519*} \\
\midrule
\multicolumn{12}{c}{\textit{Baselines}} \\
\midrule
\multicolumn{12}{l}{\textbf{BERT}} \\
\multicolumn{2}{l}{w/ Majority Vote} & 0.145 & 0.532 & 0.170 & 0.553 & 0.179 & 0.215 & 0.235 & 0.578 & 0.122 & --- \\
\multicolumn{2}{l}{w/ Indiv. Label} & 0.192 & 0.543 & 0.249 & 0.574 & 0.180 & 0.225 & 0.353 & 0.620 & 0.120 & 0.279 \\
\multicolumn{2}{l}{w/ Demo. Description} & 0.312 & \textbf{0.573} & 0.277 & \textbf{0.583} & \textbf{0.163} & \textbf{0.310} & \textbf{0.390} & \textbf{0.636} & \textbf{0.088} & \textbf{0.482} \\
\multicolumn{2}{l}{w/ Demo. Embedding} & 0.197 & 0.544 & 0.241 & 0.572 & 0.178 & 0.227 & 0.370 & 0.631 & 0.111 & 0.357 \\
\multicolumn{2}{l}{DEM-MoE} & \textbf{0.391} & 0.505 & \textbf{0.285} & 0.507 & 0.371 & 0.017 & 0.214 & 0.488 & 0.232 & 0.046 \\
\bottomrule
\end{tabular}
}
\label{tab:main-trained}
\end{table}

\section{Generalization Results}
\label{app:generalization}

We report generalization results of trained models for both cross-dataset transfer and transfer to held-out demographic groups.

\paragraph{Cross-dataset transfer} Cross-dataset transfer results of trained LLMs in Table~\ref{tab:generalization_performance} show a pattern broadly consistent with the in-domain results, with overall performance remaining comparable despite evaluation on unseen datasets.
NoDemo is still strongest at the individual level, whereas the benefits of socio-demographic modeling become clearer with aggregation. 
DemoCond and especially DemoReason improve group- and population-level performance, with DemoReason performing best on most population metrics and the continuous group-level metrics. 
This suggests that the systematic signal captured by socio-demographic modeling generalizes across datasets rather than being specific to the training distribution.

\begin{table*}[h]
\centering
\small
\caption{Cross-dataset transfer performance of trained models. 
For each source dataset and model, performance is averaged over the two target datasets as transfer targets; each cell then reports the mean across all source-dataset--model pairs.}
\label{tab:generalization_performance}
\begin{tabular}{lcccccccccc}
\toprule
\multirow{2}{*}{\textbf{Variant}}
& \multicolumn{2}{c}{\textbf{Individual}}
& \multicolumn{4}{c}{\textbf{Group}}
& \multicolumn{4}{c}{\textbf{Population}} \\
\cmidrule(lr){2-3}
\cmidrule(lr){4-7}
\cmidrule(lr){8-11}
& F1 $\uparrow$
& Acc $\uparrow$
& F1 $\uparrow$
& Acc $\uparrow$
& MSE $\downarrow$
& $\rho$ $\uparrow$
& F1 $\uparrow$
& Acc $\uparrow$
& MSE $\downarrow$
& $\rho$ $\uparrow$ \\
\midrule

NoDemo
& \textbf{0.362}
& \textbf{0.572}
& \textbf{0.363}
& 0.602
& 0.222
& 0.247
& 0.391
& 0.642
& 0.150
& 0.387 \\

DemoCond
& 0.340
& 0.566
& 0.354
& \textbf{0.604}
& 0.199
& 0.259
& 0.381
& 0.642
& 0.108
& 0.466 \\

DemoReason
& 0.345
& 0.561
& 0.361
& 0.600
& \textbf{0.198}
& \textbf{0.261}
& \textbf{0.394}
& \textbf{0.659}
& \textbf{0.088}
& \textbf{0.484} \\

\bottomrule
\end{tabular}
\end{table*}

\paragraph{Held-out demographic-group transfer} 
To evaluate generalization to unseen demographic groups, we train a Llama-8B DemoReason variant on D3CODE while holding out one group from each demographic axis (region=Oceania, country=Brazil, age=50+, gender=Other, ses=8).
We then evaluate the model separately on demographic groups seen during training and the held-out groups.
Table~\ref{tab:ood-generalization} shows that DemoReason performs comparably across seen and unseen groups overall.
On individual- and group-level discrete metrics, it even performs better on the held-out groups, whereas performance is lower for unseen groups at the population level.
The continuous metrics reveal a clearer generalization gap, with DemoReason achieving better calibration on seen than unseen groups.
Overall, these results suggest that DemoReason can generalize to unseen demographic groups.

\begin{table*}[t]
\centering
\tiny
\caption{Unseen demographic-group transfer results of Llama-8B DemoReason on D3CODE. We report both ID results, where demographic groups are observed during training, and OOD results, where the evaluated groups are held out and unseen during training.}
\label{tab:ood-generalization}
\resizebox{\textwidth}{!}{%
\begin{tabular}{llcccccccccc}
\toprule
\multirow{2}{*}{\textbf{Variant}} & \multirow{2}{*}{\textbf{Split}} & \multicolumn{2}{c}{\textbf{Individual}} & \multicolumn{4}{c}{\textbf{Group}} & \multicolumn{4}{c}{\textbf{Population}} \\
\cmidrule(lr){3-4}
\cmidrule(lr){5-8}
\cmidrule(lr){9-12}
 & & F1 $\uparrow$ & Acc $\uparrow$ & F1 $\uparrow$ & Acc $\uparrow$ & MSE $\downarrow$ & $\rho$ $\uparrow$ & F1 $\uparrow$ & Acc $\uparrow$ & MSE $\downarrow$ & $\rho$ $\uparrow$ \\
\midrule
\multirow{2}{*}{DemoReason} 
 & ID & 0.349 & 0.546 & 0.327 & 0.570 & \textbf{0.255} & \textbf{0.179} & \textbf{0.408} & \textbf{0.632} & \textbf{0.070} & \textbf{0.482} \\
 & OOD & \textbf{0.352} & \textbf{0.556} & \textbf{0.342} & \textbf{0.572} & 0.310 & 0.157 & 0.397 & 0.628 & 0.122 & 0.368 \\
\bottomrule
\end{tabular}%
}
\end{table*}

\section{Qualitative Comparison}
\label{app:qualitative}

We provide a qualitative comparison of NoDemo, DemoCond, and DemoReason in estimating the \texttt{crowd} baseline, \texttt{shift} deviation, and final judgment.
We select a representative example from D3CODE using outputs from the trained Qwen-9B variants.
The empirical crowd baseline is 0.4306, while the individual judgment is positive (offensive).
Accordingly, the individual--majority deviation proxy is positive, since the individual judgment differs from the binarized majority judgment implied by the empirical baseline.

\begin{figure}[h]
    \centering
    \includegraphics[width=\textwidth]{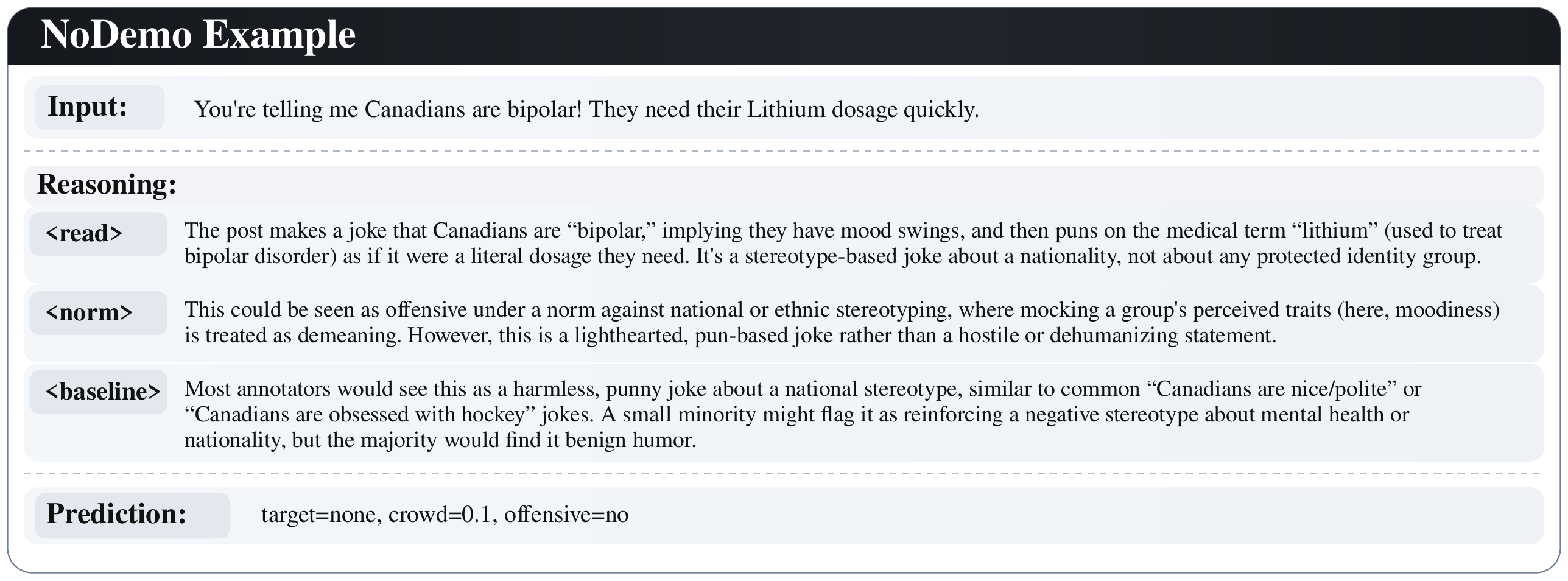}
    \caption{NoDemo example from a trained Qwen-9B model on D3CODE. Without access to the annotator's socio-demographic profile, NoDemo substantially underestimates the crowd baseline (0.1 vs. 0.4306) and incorrectly predicts the judgment as non-offensive.}
    \label{fig:nodemo_example}
\end{figure}

\begin{figure}[h]
    \centering
    \includegraphics[width=\textwidth]{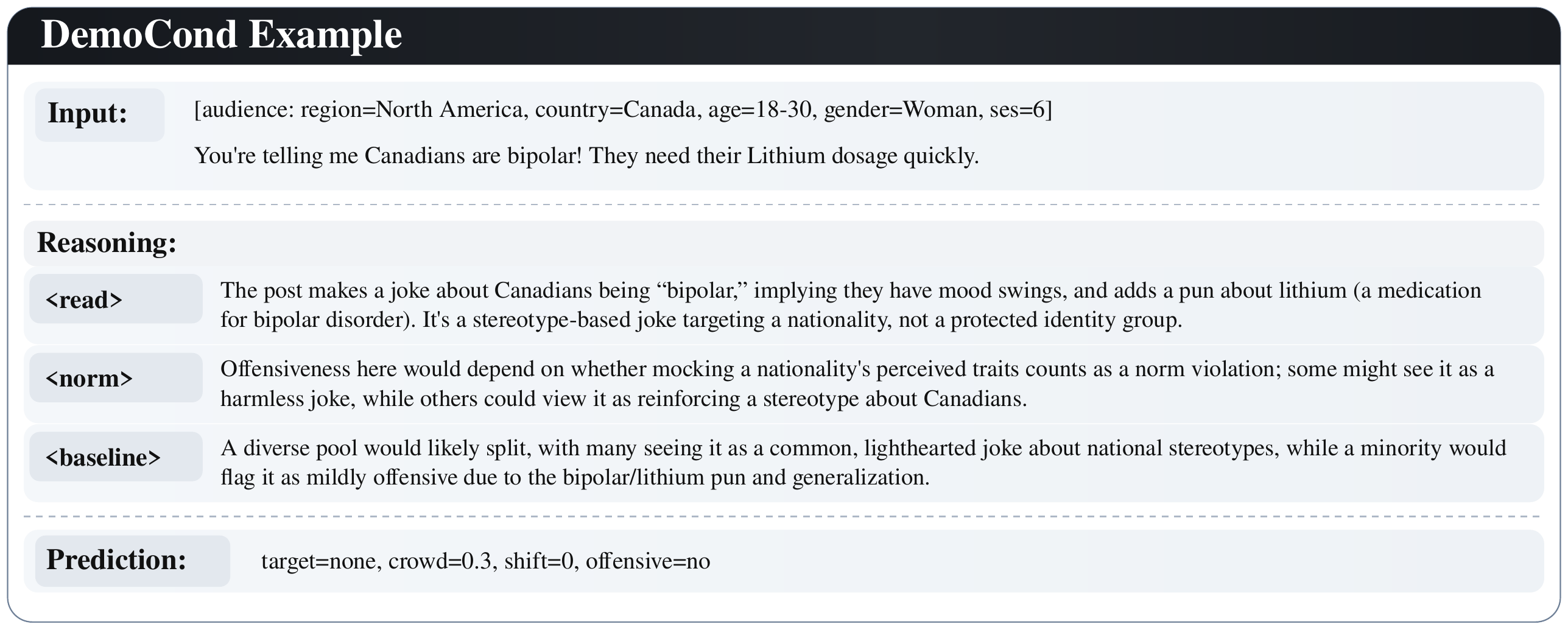}
    \caption{DemoCond example from a trained Qwen-9B model on D3CODE. Compared with NoDemo, DemoCond recognizes greater disagreement in how annotators may perceive the post and therefore provides a more accurate crowd baseline estimate (0.3 vs. 0.4306). However, without explicitly reasoning about the perception of the specific annotator, DemoCond predicts \texttt{shift}=0, fails to capture the individual--majority deviation, and still makes an incorrect final prediction.}
    \label{fig:democond_example}
\end{figure}

\begin{figure}[h]
    \centering
    \includegraphics[width=\textwidth]{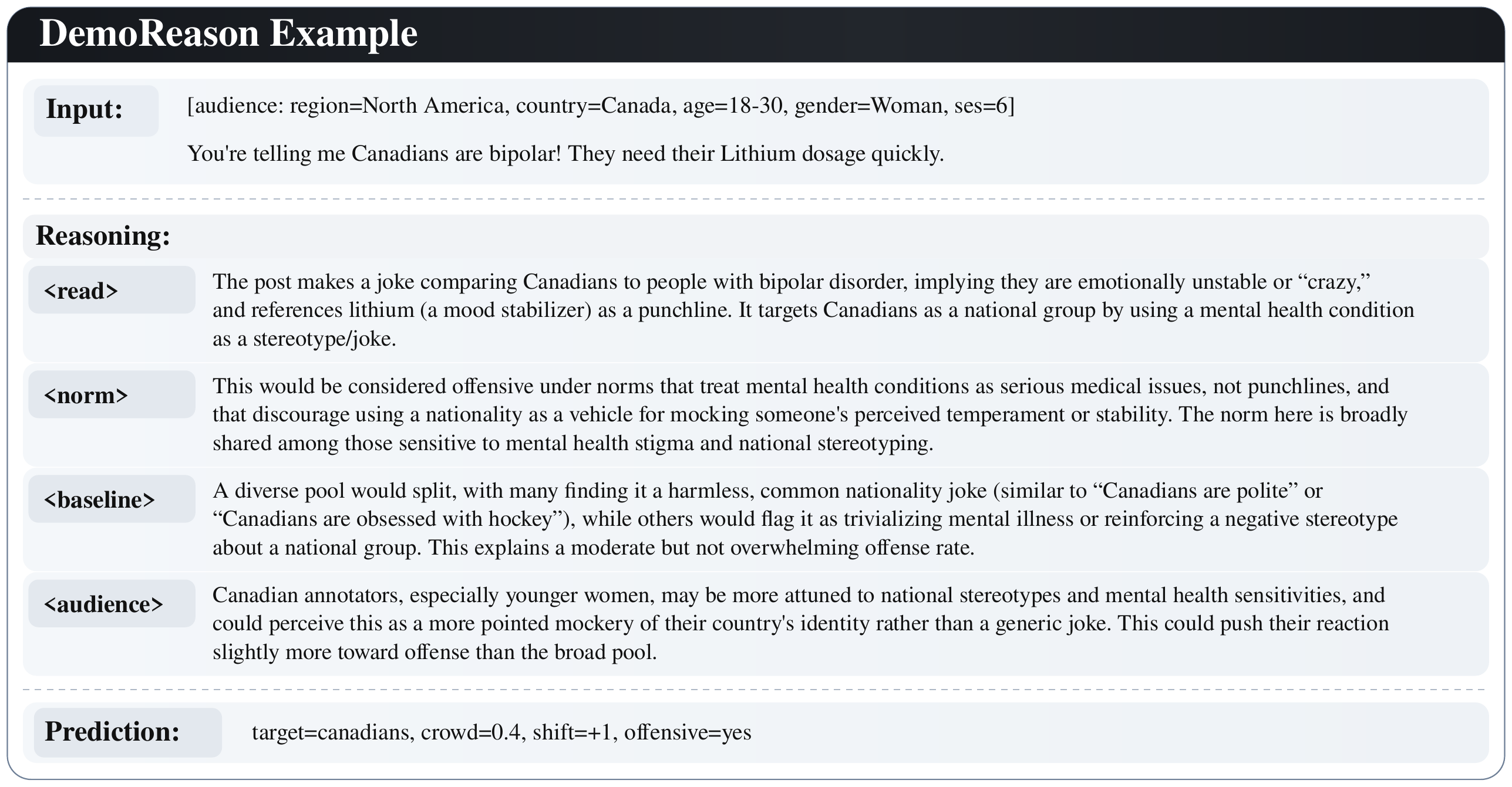}
    \caption{DemoReason example from a trained Qwen-9B model on D3CODE. Compared with DemoCond, DemoReason provides a more accurate \texttt{crowd} estimate (0.4 vs. 0.4306). With explicit socio-demographic reasoning, DemoReason identifies that a Canadian annotator may be more sensitive to a post targeting Canadians and therefore predicts \texttt{shift}=+1, correctly capturing the individual--majority deviation and yielding the correct offensive final prediction.}
    \label{fig:demoreason_example}
\end{figure}

Figures~\ref{fig:nodemo_example},~\ref{fig:democond_example} and~\ref{fig:demoreason_example} illustrate how the three variants differ in estimating the \texttt{crowd} baseline and annotator-specific \texttt{shift}.
Without access to the annotator's socio-demographic profile, NoDemo substantially underestimates the crowd baseline (0.1 vs. the empirical 0.4306) and predicts the judgment as non-offensive.
Conditioning on socio-demographics, DemoCond recognizes greater disagreement in how the post may be perceived and produces a more accurate baseline estimate of 0.3. However, without explicitly reasoning about how the specific annotator may deviate from the crowd, it predicts \texttt{shift}=0 and still yields an incorrect non-offensive prediction. DemoReason further improves the baseline estimate to 0.4 and, through explicit socio-demographic reasoning, identifies that a Canadian annotator may be more sensitive to a post targeting Canadians. It consequently predicts \texttt{shift}=+1, correctly capturing the individual--majority deviation and yielding the correct offensive judgment.
This example illustrates the mechanisms identified in Section~\ref{sec:where}: incorporating socio-demographics improves baseline estimation, with explicit reasoning yielding a more accurate estimate than demographic conditioning alone. Explicit reasoning also captures a meaningful demographic-specific deviation, whereas simple conditioning fails to recover this shift. Together, the improved estimates of both components lead to a more accurate final prediction.
However, the reasoning also reveals a potential risk: DemoReason may rely on stereotypical assumptions about demographic groups, such as suggesting that women are more sensitive to certain forms of offensiveness.

\section{Full Baseline Estimation Results}
\label{app:full_baseline}

Tables~\ref{tab:crowd-prompted} and~\ref{tab:crowd-trained} report per-model annotation- and item-level baseline estimation results. 
Incorporating socio-demographics generally improves baseline estimation in the baseline models as well.

\begin{table}[t]
\centering
\small
\caption{
        Annotation- and item-level baseline estimation results for all \textit{prompted} models and baselines.
        Metric scores are averaged over three datasets, and \textit{Overall} averages across backbones.
        * denotes significant improvement over NoDemo; ** denotes significant improvement over both NoDemo and the other variant across model--dataset pairs (paired $t$-test, $p<0.05$).
        }
\resizebox{\textwidth}{!}{%
\begin{tabular}{llcccccccc}
\toprule
\multirow{2}{*}{\textbf{Model}} & \multirow{2}{*}{\textbf{Variant}} & \multicolumn{4}{c}{\textbf{Baseline Estimation (Ann.)}} & \multicolumn{4}{c}{\textbf{Baseline Estimation (Item)}} \\
\cmidrule(lr){3-6}
\cmidrule(lr){7-10}
 &  & MSE $\downarrow$ & MAE $\downarrow$ & $\rho \uparrow$ & W1 $\downarrow$ & MSE $\downarrow$ & MAE $\downarrow$ & $\rho \uparrow$ & W1 $\downarrow$ \\
\midrule
\multirow{3}{*}{Llama-3B} & NoDemo & \textbf{0.060} & \textbf{0.186} & \textbf{0.320} & 0.098 & 0.059 & 0.185 & 0.333 & \textbf{0.097} \\
 & DemoCond & 0.060 & 0.193 & 0.286 & 0.098 & \textbf{0.041} & \textbf{0.161} & \textbf{0.501} & 0.112 \\
 & DemoReason & 0.068 & 0.204 & 0.236 & \textbf{0.078} & 0.044 & 0.168 & 0.394 & 0.098 \\
\midrule
\multirow{3}{*}{Qwen-4B} & NoDemo & 0.093 & 0.238 & \textbf{0.555} & 0.155 & 0.089 & 0.234 & 0.565 & 0.154 \\
 & DemoCond & 0.083 & 0.223 & 0.516 & 0.129 & 0.066 & 0.198 & 0.588 & 0.113 \\
 & DemoReason & \textbf{0.076} & \textbf{0.210} & 0.552 & \textbf{0.117} & \textbf{0.061} & \textbf{0.187} & \textbf{0.619} & \textbf{0.103} \\
\midrule
\multirow{3}{*}{Llama-8B} & NoDemo & 0.049 & 0.167 & 0.468 & 0.077 & 0.047 & 0.164 & 0.501 & 0.079 \\
 & DemoCond & \textbf{0.048} & \textbf{0.167} & 0.516 & \textbf{0.074} & \textbf{0.038} & \textbf{0.152} & 0.617 & \textbf{0.076} \\
 & DemoReason & 0.050 & 0.175 & \textbf{0.565} & 0.092 & 0.044 & 0.167 & \textbf{0.620} & 0.095 \\
\midrule
\multirow{3}{*}{Qwen-9B} & NoDemo & 0.110 & 0.266 & 0.442 & 0.190 & 0.111 & 0.269 & 0.457 & 0.195 \\
 & DemoCond & 0.091 & 0.236 & 0.509 & 0.154 & 0.078 & 0.217 & 0.597 & 0.143 \\
 & DemoReason & \textbf{0.082} & \textbf{0.219} & \textbf{0.533} & \textbf{0.134} & \textbf{0.067} & \textbf{0.197} & \textbf{0.614} & \textbf{0.117} \\
\midrule
\multirow{3}{*}{Qwen-27B} & NoDemo & 0.081 & 0.220 & \textbf{0.602} & 0.145 & 0.078 & 0.217 & 0.622 & 0.144 \\
 & DemoCond & \textbf{0.081} & \textbf{0.217} & 0.597 & \textbf{0.142} & \textbf{0.073} & \textbf{0.208} & 0.641 & 0.137 \\
 & DemoReason & 0.082 & 0.220 & \textbf{0.602} & 0.142 & 0.074 & 0.210 & \textbf{0.652} & \textbf{0.137} \\
\midrule
\multirow{3}{*}{Llama-70B} & NoDemo & 0.045 & 0.157 & \textbf{0.601} & \textbf{0.059} & 0.042 & 0.154 & 0.613 & \textbf{0.063} \\
 & DemoCond & 0.041 & 0.153 & \textbf{0.601} & 0.072 & 0.037 & 0.147 & \textbf{0.634} & 0.074 \\
 & DemoReason & \textbf{0.039} & \textbf{0.149} & 0.584 & 0.069 & \textbf{0.035} & \textbf{0.143} & 0.625 & 0.068 \\
\midrule
\multirow{3}{*}{Qwen-122B} & NoDemo & 0.064 & 0.187 & 0.541 & 0.124 & 0.061 & 0.184 & 0.570 & 0.122 \\
 & DemoCond & 0.056 & 0.175 & 0.583 & 0.103 & 0.050 & 0.165 & 0.623 & 0.102 \\
 & DemoReason & \textbf{0.052} & \textbf{0.167} & \textbf{0.588} & \textbf{0.087} & \textbf{0.046} & \textbf{0.159} & \textbf{0.626} & \textbf{0.087} \\
\midrule
\multirow{3}{*}{GPT-4o-mini} & NoDemo & 0.051 & 0.179 & \textbf{0.594} & 0.109 & 0.050 & 0.179 & 0.621 & 0.112 \\
 & DemoCond & 0.044 & 0.157 & 0.580 & \textbf{0.083} & 0.038 & 0.151 & 0.627 & \textbf{0.085} \\
 & DemoReason & \textbf{0.039} & \textbf{0.151} & 0.592 & 0.085 & \textbf{0.035} & \textbf{0.146} & \textbf{0.641} & 0.090 \\
\midrule
\multirow{3}{*}{Haiku-4.5} & NoDemo & 0.047 & 0.156 & 0.589 & 0.069 & 0.046 & 0.158 & 0.594 & 0.073 \\
 & DemoCond & 0.044 & 0.153 & 0.576 & \textbf{0.066} & 0.040 & \textbf{0.146} & 0.602 & \textbf{0.067} \\
 & DemoReason & \textbf{0.043} & \textbf{0.153} & \textbf{0.590} & 0.075 & \textbf{0.039} & 0.148 & \textbf{0.614} & 0.079 \\
\midrule
\multirow{3}{*}{Overall} & NoDemo & 0.067 & 0.195 & 0.524 & 0.114 & 0.065 & 0.194 & 0.542 & 0.115 \\
 & DemoCond & 0.061 & 0.186 & 0.529 & 0.102* & 0.051* & 0.172* & \textbf{0.603*} & 0.101* \\
 & DemoReason & \textbf{0.059} & \textbf{0.183*} & \textbf{0.538} & \textbf{0.097*} & \textbf{0.050*} & \textbf{0.169*} & 0.601* & \textbf{0.097} \\
\midrule
\multicolumn{10}{c}{\textit{Baselines}} \\
\midrule
\multicolumn{10}{l}{\textbf{GPT-4o-mini}} \\
\multicolumn{2}{l}{Zero-shot} & 0.067 & 0.197 & \textbf{0.449} & 0.105 & 0.069 & 0.203 & 0.470 & 0.113 \\
\multicolumn{2}{l}{Zero-shot + Demo.} & \textbf{0.063} & \textbf{0.192} & 0.428 & \textbf{0.095} & 0.047 & 0.167 & 0.575 & \textbf{0.083} \\
\multicolumn{2}{l}{CoT} & 0.084 & 0.221 & 0.402 & 0.114 & 0.083 & 0.220 & 0.402 & 0.116 \\
\multicolumn{2}{l}{CoT + Demo.} & 0.090 & 0.225 & 0.371 & 0.112 & \textbf{0.040} & \textbf{0.163} & \textbf{0.587} & 0.106 \\
\midrule
\multicolumn{10}{l}{\textbf{Haiku-4.5}} \\
\multicolumn{2}{l}{Zero-shot} & 0.054 & 0.171 & 0.531 & 0.137 & 0.052 & 0.169 & 0.545 & 0.132 \\
\multicolumn{2}{l}{Zero-shot + Demo.} & \textbf{0.051} & \textbf{0.168} & 0.545 & 0.117 & 0.045 & 0.157 & 0.600 & 0.095 \\
\multicolumn{2}{l}{CoT} & 0.054 & 0.175 & 0.509 & \textbf{0.094} & 0.053 & 0.173 & 0.509 & 0.094 \\
\multicolumn{2}{l}{CoT + Demo.} & 0.053 & 0.176 & \textbf{0.546} & 0.100 & \textbf{0.040} & \textbf{0.152} & \textbf{0.616} & \textbf{0.071} \\
\bottomrule
\end{tabular}
}
\label{tab:crowd-prompted}
\end{table}

\begin{table}[h]
\centering
\small
\caption{
        Annotation- and item-level baseline estimation results for all \textit{trained} models and baselines.
        Metric scores are averaged over three datasets, and \textit{Overall} averages across backbones.
        * denotes significant improvement over NoDemo; ** denotes significant improvement over both NoDemo and the other variant across model--dataset pairs (paired $t$-test, $p<0.05$).
        }
\resizebox{\textwidth}{!}{%
\begin{tabular}{llcccccccc}
\toprule
\multirow{2}{*}{\textbf{Model}} & \multirow{2}{*}{\textbf{Variant}} & \multicolumn{4}{c}{\textbf{Baseline Estimation (Ann.)}} & \multicolumn{4}{c}{\textbf{Baseline Estimation (Item)}} \\
\cmidrule(lr){3-6}
\cmidrule(lr){7-10}
 &  & MSE $\downarrow$ & MAE $\downarrow$ & $\rho \uparrow$ & W1 $\downarrow$ & MSE $\downarrow$ & MAE $\downarrow$ & $\rho \uparrow$ & W1 $\downarrow$ \\
\midrule
\multirow{3}{*}{Llama-3B} & NoDemo & 0.063 & 0.198 & 0.424 & 0.135 & 0.058 & 0.189 & 0.485 & 0.133 \\
 & DemoCond & \textbf{0.054} & 0.184 & \textbf{0.452} & 0.116 & 0.044 & 0.171 & 0.545 & 0.115 \\
 & DemoReason & 0.054 & \textbf{0.180} & 0.440 & \textbf{0.089} & \textbf{0.041} & \textbf{0.158} & \textbf{0.565} & \textbf{0.093} \\
\midrule
\multirow{3}{*}{Qwen-4B} & NoDemo & \textbf{0.039} & \textbf{0.144} & \textbf{0.594} & \textbf{0.055} & 0.034 & 0.137 & \textbf{0.653} & \textbf{0.046} \\
 & DemoCond & 0.041 & 0.151 & 0.559 & 0.056 & \textbf{0.033} & \textbf{0.135} & 0.640 & 0.051 \\
 & DemoReason & 0.042 & 0.153 & 0.559 & 0.066 & 0.033 & 0.138 & 0.643 & 0.057 \\
\midrule
\multirow{3}{*}{Llama-8B} & NoDemo & 0.044 & 0.157 & \textbf{0.515} & 0.072 & 0.039 & 0.148 & 0.586 & 0.072 \\
 & DemoCond & 0.044 & 0.155 & 0.508 & 0.080 & 0.034 & 0.138 & \textbf{0.601} & 0.074 \\
 & DemoReason & \textbf{0.040} & \textbf{0.150} & 0.506 & \textbf{0.067} & \textbf{0.032} & \textbf{0.133} & 0.575 & \textbf{0.071} \\
\midrule
\multirow{3}{*}{Qwen-9B} & NoDemo & 0.054 & 0.175 & 0.488 & 0.131 & 0.049 & 0.167 & 0.541 & 0.124 \\
 & DemoCond & \textbf{0.049} & \textbf{0.164} & \textbf{0.522} & \textbf{0.107} & \textbf{0.041} & \textbf{0.150} & \textbf{0.628} & 0.102 \\
 & DemoReason & 0.051 & 0.170 & 0.501 & 0.109 & 0.042 & 0.153 & 0.614 & \textbf{0.099} \\
\midrule
\multirow{3}{*}{Overall} & NoDemo & 0.050 & 0.168 & 0.505 & 0.098 & 0.045 & 0.160 & 0.566 & 0.094 \\
 & DemoCond & 0.047 & 0.164 & \textbf{0.510} & 0.090 & 0.038* & 0.148* & \textbf{0.604*} & 0.086 \\
 & DemoReason & \textbf{0.047} & \textbf{0.163} & 0.502 & \textbf{0.083**} & \textbf{0.037*} & \textbf{0.146*} & 0.599* & \textbf{0.080*} \\
\midrule
\multicolumn{10}{c}{\textit{Baselines}} \\
\midrule
\multicolumn{10}{l}{\textbf{BERT}} \\
\multicolumn{2}{l}{w/ Majority Vote} & 0.059 & 0.189 & 0.522 & 0.146 & 0.059 & 0.187 & 0.541 & 0.142 \\
\multicolumn{2}{l}{w/ Indiv. Label} & 0.030 & \textbf{0.123} & \textbf{0.632} & 0.045 & 0.029 & \textbf{0.122} & 0.647 & \textbf{0.045} \\
\multicolumn{2}{l}{w/ Demo. Description} & 0.040 & 0.147 & 0.529 & \textbf{0.042} & 0.030 & 0.123 & \textbf{0.650} & 0.048 \\
\multicolumn{2}{l}{w/ Demo. Embedding} & \textbf{0.029} & 0.124 & 0.625 & 0.043 & \textbf{0.028} & 0.122 & \textbf{0.650} & 0.046 \\
\multicolumn{2}{l}{DEM-MoE} & 0.091 & 0.266 & 0.017 & 0.260 & 0.090 & 0.265 & 0.027 & 0.262 \\
\bottomrule
\end{tabular}
}
\label{tab:crowd-trained}
\end{table}

\section{Baseline- and Shift-only Intervention Tests}
\label{app:intervention_test}

Beyond the joint intervention on \texttt{crowd} and \texttt{shift}, Tables~\ref{tab:intervention-crowdonly} and~\ref{tab:intervention-shiftonly} report separate interventions on the two components.
The baseline-only intervention more consistently follows the expected Oracle $>$ Original $>$ Adversarial ordering, providing stronger evidence that baseline quality directly affects final predictions.
In contrast, this ordering is less consistent for shift-only interventions, where a common failure mode is that adversarial shifts outperform the original ones.
This is consistent with our earlier finding that model-emitted \texttt{shift} captures only modest effect sizes.

\begin{table*}[t]
\centering
\small
\caption{Performance under baseline-only interventions on baseline and shift estimates for Llama-8B (trained) and Llama-70B (prompted). 
Each cell reports \textit{Original / Oracle / Adversarial} performance.
Bold marks cells following the expected ordering Oracle $>$ Original $>$ Adversarial.}
\label{tab:intervention-crowdonly}
\resizebox{\textwidth}{!}{%
\begin{tabular}{llccccc}
\toprule
\multirow{2}{*}{\textbf{Model}} & \multirow{2}{*}{\textbf{Variant}} & \textbf{Individual} & \multicolumn{2}{c}{\textbf{Group}} & \multicolumn{2}{c}{\textbf{Population}} \\
\cmidrule(lr){3-3}
\cmidrule(lr){4-5}
\cmidrule(lr){6-7}
 & & F1 $\uparrow$ & F1 $\uparrow$ & MSE $\downarrow$ & F1 $\uparrow$ & MSE $\downarrow$ \\
\midrule
        \multirow{2}{*}{Llama-8B (Trained)} & DemoCond & 0.453/0.441/0.337 & \textbf{0.442/0.477/0.410} & 0.190/0.152/0.168 & \textbf{0.549/0.650/0.457} & \textbf{0.094/0.081/0.096} \\
         & DemoReason & \textbf{0.474/0.478/0.443} & \textbf{0.450/0.453/0.421} & \textbf{0.188/0.185/0.197} & 0.526/0.595/0.556 & \textbf{0.059/0.059/0.063} \\
         \midrule
        \multirow{2}{*}{Llama-70B (Prompted)} & DemoCond & \textbf{0.438/0.510/0.335} & \textbf{0.425/0.511/0.270} & \textbf{0.178/0.164/0.195} & \textbf{0.435/0.667/0.353} & 0.095/0.075/0.078 \\
         & DemoReason & \textbf{0.458/0.549/0.432} & \textbf{0.443/0.534/0.413} & 0.180/0.186/0.196 & \textbf{0.500/0.632/0.468} & 0.084/0.077/0.073 \\
\bottomrule
\end{tabular}%
}
\end{table*}

\begin{table*}[t]
\centering
\small
\caption{Performance under shift-only interventions on baseline and shift estimates for Llama-8B (trained) and Llama-70B (prompted). 
Each cell reports \textit{Original / Oracle / Adversarial} performance.
Bold marks cells following the expected ordering Oracle $>$ Original $>$ Adversarial.}
\label{tab:intervention-shiftonly}
\resizebox{\textwidth}{!}{%
\begin{tabular}{llccccc}
\toprule
\multirow{2}{*}{\textbf{Model}} & \multirow{2}{*}{\textbf{Variant}} & \textbf{Individual} & \multicolumn{2}{c}{\textbf{Group}} & \multicolumn{2}{c}{\textbf{Population}} \\
\cmidrule(lr){3-3}
\cmidrule(lr){4-5}
\cmidrule(lr){6-7}
 & & F1 $\uparrow$ & F1 $\uparrow$ & MSE $\downarrow$ & F1 $\uparrow$ & MSE $\downarrow$ \\
\midrule
        \multirow{2}{*}{Llama-8B (Trained)} & DemoCond & 0.453/0.412/0.333 & \textbf{0.442/0.483/0.398} & 0.190/0.131/0.165 & 0.549/0.378/0.432 & 0.094/0.110/0.108 \\
         & DemoReason & 0.474/0.582/0.492 & 0.450/0.527/0.450 & \textbf{0.188/0.149/0.200} & 0.526/0.500/0.600 & 0.059/0.075/0.071 \\
         \midrule
        \multirow{2}{*}{Llama-70B (Prompted)} & DemoCond & 0.438/0.538/0.499 & 0.425/0.495/0.463 & \textbf{0.178/0.167/0.183} & 0.435/0.480/0.519 & 0.095/0.095/0.088 \\
         & DemoReason & 0.458/0.525/0.500 & 0.443/0.495/0.482 & \textbf{0.180/0.175/0.188} & 0.500/0.528/0.538 & \textbf{0.084/0.086/0.084} \\
\bottomrule
\end{tabular}%
}
\end{table*}

\section{Limitations}
\label{app:limitations}
First, some of our analyses, particularly those involving demographic-specific \texttt{shift}, rely on empirical or proxy targets because the underlying quantities are not directly observable.
These proxies are inevitably affected by annotation noise and may therefore underestimate or distort the true signal.
Moreover, the empirical baseline reflects the annotator population available in each dataset and may not perfectly represent a broader population.
To mitigate this issue, we evaluate the relevant effects at multiple levels of aggregation and focus on datasets or annotation pools with sufficiently large numbers of annotations.
Nevertheless, some findings should still be interpreted with these measurement and sampling limitations in mind.

Second, our findings are specific to the prompting templates, reasoning format, and training pipeline considered in this work.
Although we observe consistent patterns across multiple datasets and models, and report performance of several baseline models in Appendices~\ref{app:full_performance} and~\ref{app:full_baseline}, it remains unclear whether they generalize to other ways of incorporating socio-demographic information, other subjective tasks beyond offensiveness, or different annotator populations.
Furthermore, although our format explicitly elicits \texttt{crowd} baseline and \texttt{shift} estimates, LLMs are not constrained to derive their final predictions from these components.
Our intervention results establish their functional relevance to prediction, but do not demonstrate that the model internally follows the assumed decomposition.

Finally, the available socio-demographic attributes provide only a coarse representation of annotator identity and the factors underlying subjective judgments.
Our analyses primarily consider groups defined along individual demographic axes, while intersectional effects are difficult to estimate reliably because of data sparsity.
Socio-demographics also explain only a limited portion of individual variation, leaving many potentially relevant individual, contextual, and experiential factors unobserved.
\end{document}